\documentclass[journal]{IEEEtran}

\usepackage{array}
\usepackage{cite}

\usepackage{algorithm}
\usepackage{algorithmic}
\usepackage{mdwmath}

\usepackage{graphicx}        
\usepackage{multicol}        
\usepackage{subfigure}
\usepackage{pifont}

\usepackage{bm}
\usepackage{comment}
\usepackage{amsmath}
\usepackage{enumerate}  
\usepackage{mathrsfs}
\usepackage{makeidx}         
\usepackage{graphicx}        
\usepackage{multicol}        
\usepackage{subfigure}
\usepackage{epsfig,amssymb,latexsym}
\usepackage{psfrag}
\usepackage{algorithmic}
\usepackage{algorithm}

\usepackage{fancyhdr}
\usepackage{multirow}

\usepackage{color}
\usepackage{indentfirst}

\renewcommand{\algorithmicrequire}{\textbf{Input:}}   
\renewcommand{\algorithmicensure}{\textbf{Output:}}   

\newcommand{\mr}{\mathrm}
\newcommand{\mc}{\mathcal}

\usepackage{environ}
\usepackage{tikz}
\NewEnviron{elaboration}{
	\par
	\begin{tikzpicture}
	\node[rectangle,minimum width=0.45\textwidth] (m) {\begin{minipage}{0.26\textwidth}\BODY\end{minipage}};
	\draw[dashed] (m.south west) rectangle (m.north east);
	\end{tikzpicture}
}

\begin{document}

\title{Evolving Metric Learning for Incremental and Decremental Features}

\author{
	Jiahua~Dong,
	Yang~Cong,~\IEEEmembership{Senior~Member,~IEEE,}
	Gan~Sun,
	Tao~Zhang,
	Xu~Tang
	and Xiaowei~Xu 
	
	\thanks{This work is supported by the National Key Research and Development Program of China (2019YFB1310300) and National Nature Science Foundation of China under Grant (61722311, 61821005, 62003336). \textit{(Corresponding author: Yang Cong.)}}
	
	\thanks{Jiahua Dong and Tao Zhang are with the State Key Laboratory of Robotics, Shenyang Institute of Automation, Chinese Academy of Sciences, Shenyang 110016, China, and also with the Institutes for Robotics and Intelligent Manufacturing, Chinese Academy of Sciences, Shenyang 110016, China, and also with the University of Chinese Academy of Sciences, Beijing 100049, China (e-mail: dongjiahua1995@gmail.com, zhangtao2@sia.cn).}
	
	\thanks{Yang Cong, Gan Sun and Xu Tang are with the State Key Laboratory of Robotics, Shenyang Institute of Automation, Chinese Academy of Sciences, Shenyang 110016, China, and also with the Institutes for Robotics and Intelligent Manufacturing, Chinese Academy of Sciences, Shenyang 110016, China (e-mail: congyang81@gmail.com, sungan1412@gmail.com, tangxu@sia.cn).}
	
	\thanks{Xiaowei Xu is with the Department of Information Science, University of Arkansas at Little Rock, Arkansas 72204, USA (e-mail: xwxu@ualr.edu).}
	
}

\markboth{IEEE TRANSACTIONS ON CIRCUITS AND SYSTEMS FOR VIDEO TECHNOLOGY,~Vol.~14, No.~8, June~2021}%
{Shell \MakeLowercase{\textit{et al.}}: Bare Demo of IEEEtran.cls for IEEE Journals}

\maketitle

\begin{abstract}
Online metric learning has been widely exploited for large-scale data classification due to the low computational cost. However, amongst online practical scenarios where the features are evolving (\emph{e.g.}, some features are vanished and some new features are augmented), most metric learning models cannot be successfully applied to these scenarios, although they can tackle the evolving instances efficiently. To address the challenge, we develop a new online \underline{E}volving \underline{M}etric \underline{L}earning (EML) model for incremental and decremental features, which can handle the instance and feature evolutions simultaneously by incorporating with a smoothed Wasserstein metric distance. Specifically, our model contains two essential stages: a Transforming stage (T-stage) and a Inheriting stage (I-stage). For the T-stage, we propose to extract important information from vanished features while neglecting non-informative knowledge, and forward it into survived features by transforming them into a low-rank discriminative metric space. It further explores the intrinsic low-rank structure of heterogeneous samples to reduce the computation and memory burden especially for highly-dimensional large-scale data. 
For the I-stage, we inherit the metric performance of survived features from the T-stage and then expand to include the new augmented features. Moreover, a smoothed Wasserstein distance is utilized to characterize the similarity relationships among the heterogeneous and complex samples, since the evolving features are not strictly aligned in the different stages. In addition to tackling the challenges in one-shot case, we also extend our model into multi-shot scenario. 
After deriving an efficient optimization strategy for both T-stage and I-stage, extensive experiments on several datasets verify the superior performance of our EML model.

\end{abstract}


 \ifCLASSOPTIONpeerreview
 \begin{center} \bfseries EDICS Category: 3-BBND \end{center}
 \fi
%
\IEEEpeerreviewmaketitle

\section{Introduction}
\IEEEPARstart{M}etric learning has been successfully extended into many fields, \emph{e.g.},  face identification \cite{10.1007/978-3-642-19309-5_55}, object recognition \cite{Xu:2018:BDM:3327144.3327333} and medical diagnosis \cite{Boukouvalas2011DistanceML}. To efficiently solve the large-scale streaming data problem, learning an online discriminative metric (\emph{i.e.}, online metric learning \cite{Chechik:2010:LSO:1756006.1756042,NIPS2008_3446}) attracts lots of appealing attentions. Generally, most online metric learning models pay attention to the fast metric updating mechanisms \cite{Weinberger2009, LI2018302, NIPS2009_3703, NIPS2011_4392} or fast similarity searching strategies \cite{NIPS2008_3446, Davis2007IML1273496, NIPS2009_3703} for large-scale streaming data, where the streaming data indicate the continuous data flow that the data samples arrive consecutively in a real-time manner.

\begin{figure}[t]
	\centering
	\includegraphics[scale=0.54]{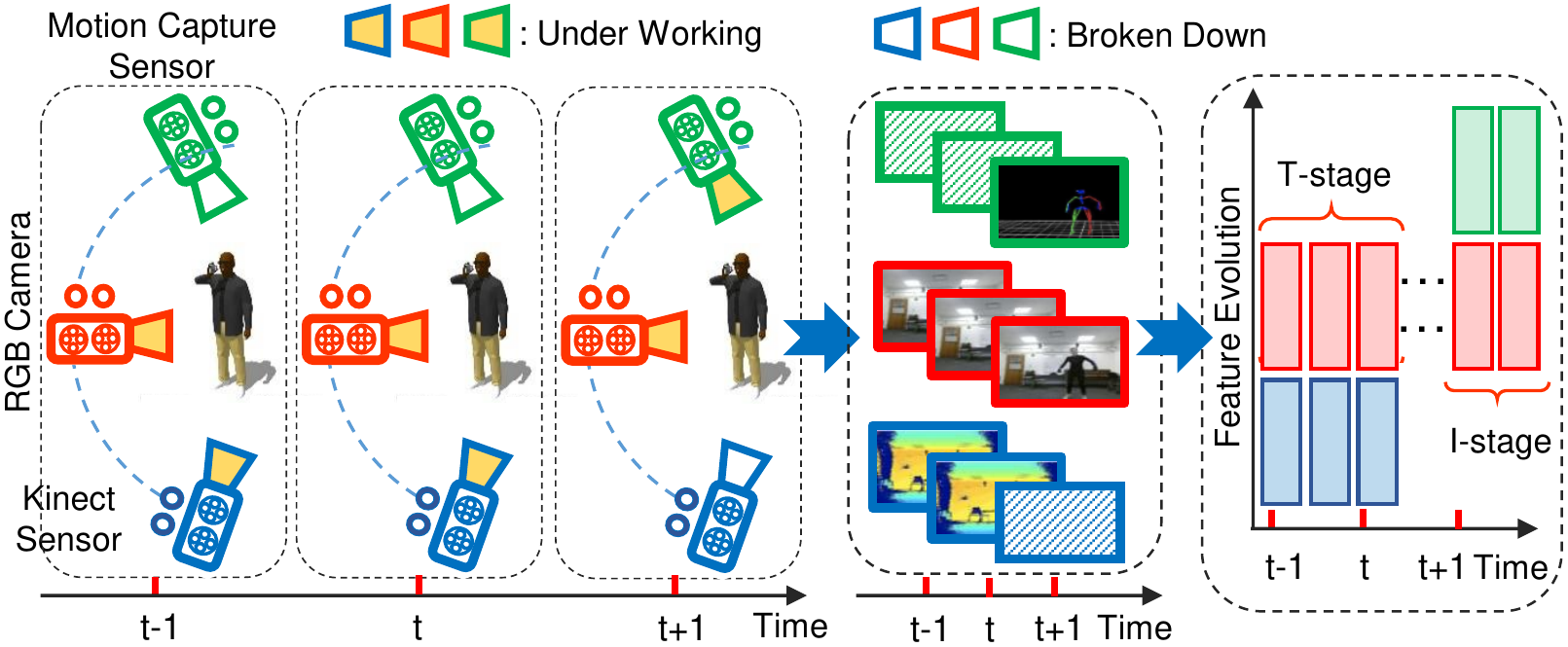}
	\vspace{-15pt}
	\caption{Illustration example of feature evolution on human motion recognition task, where the blue, red and green colors respectively indicate the vanished features collected from Kinect sensor, the survived features collected from RGB camera and the augmented features collected from motion capture senor with different lifespans. The vanished features collected from Kinect sensors are decremental in the T-stage, and the augmented features collected from from motion capture senor are incremental in the I-stage. The survived features collected from RGB camera exist in both T-stage and I-stage.}
	\label{fig:demonstration}
\end{figure}

However, these existing online metric learning methods \cite{NIPS2008_3446, Weinberger2009, Davis2007IML1273496, 7258342, 7937828} only focus on instance evolution, and ignore the feature evolution in many real-world applications, where some features are vanished and some new features are augmented. 
Take the human motion recognition \cite{DBLP:journals/corr/abs-1904-12602} as an example, as depicted in Fig.~\ref{fig:demonstration}, the sudden damage of Kinect sensor results in the absence of depth information of human motion, while the emerging of new motion capture sensor could obtain the auxiliary human skeleton knowledge for motion recognition. It leads to a corresponding decrease and increase in the feature dimensionality of the input data, which are considered as the vanished features and augmented features, respectively. The features collected from RGB camera that has been working are regarded as survived features. Such feature evolution setting heavily cripples the human motion recognition performance of the pre-trained model \cite{DBLP:journals/corr/abs-1904-12602}.
Another interesting example is that different sensors (\emph{e.g.}, radioisotope, trace metal and biological sensors \cite{s5010004}) are deployed to monitor the dynamic environment change in full aspects. Some sensors expire (vanished features) whereas some new sensors are deployed (augmented features) when different electrochemical conditions and lifespans occur. A fixed or static online metric learning model will fail to take advantage of sensors evolved in this way. Therefore, how to establish a novel metric learning model to simultaneously handle both instance and feature evolutions amongst these online practical systems is our main focus in this paper.

To address the challenges above, as illustrated in Fig.~\ref{fig:demonstration}, we develop a new online \underline{E}volving \underline{M}etric \underline{L}earning (EML) model for incremental and decremental features, which can exploit streaming data with both instance and feature evolutions in an online manner. To be specific, the proposed EML model consists of two significant stages, \emph{i.e.}, a Transforming stage (T-stage) and a Inheriting stage (I-stage). 1) In the T-stage where features are decremental, we propose to explore the important information and data structure from vanished features, and transform them into a low-rank discriminative metric space of survived features, which could be utilized to promote the learning process of the I-stage. Moreover, it explores the intrinsic low-rank structure of the streaming data, which efficiently reduces both memory and computation costs especially for large-scale samples with high dimensional feature. 
2) For the I-stage where features are incremental, based on the learned discriminative metric space in the T-stage, we inherit the metric performance of survived features from T-stage, and then expand to consider new augmented features. Furthermore, to better explore the similarity relations amongst the heterogeneous data, a smoothed Wasserstein distance is applied to both T-stage and I-stage where the evolving features are strictly unaligned and heterogeneous in different stages. For the model optimization, we derive an efficient optimization strategy to solve the formulations of T-stage and I-stage. 
Besides, our EML model could be successfully extended from one-shot scenario into multi-shot scenario, where one-shot scenario indicates that the features of streaming data would only be incremental and decremental by one time (as shown in Fig.~\ref{fig:main_algorithm}), while multi-shot scenario denotes that the representations of streaming data would be incremental and decremental multiple times (as shown in Fig.~\ref{fig:multi-shot}). 
Comprehensive experimental results on several datasets strongly support the effectiveness of our proposed EML model.

The main contributions of this paper are summarized as follows:
\begin{itemize}
	\setlength{\itemsep}{0.5pt}
	\setlength{\parsep}{0pt}
	\setlength{\parskip}{0pt}
	\item We propose an online Evolving Metric Learning (EML) model for incremental and decremental features to tackle both instance and feature evolutions simultaneously. To our best knowledge, this is the first exploration to tackle this crucial, but rarely-researched challenge in the metric learning field.  
		
	\item We present two stages for both feature and instance evolutions, \emph{i.e.}, a Transforming stage (T-stage) and a Inheriting stage (I-stage), which can not only make full use of the vanished features in the T-stage, but also take advantage of streaming data with new augmented features in the I-stage. 
	
	\item A smoothed Wasserstein distance is incorporated into metric learning to characterize the similarity relations of heterogeneous evolving features among different stages. After deriving an alternating direction optimization algorithm to optimize our EML model, extensive experiments on representative datasets validate the superior performance of our proposed EML model.
	
\end{itemize}

\section{Related Work}\label{sec:related work}
This section provides a brief overview about metric learning, followed by some representative methods about feature evolution.

\subsection{Metric Learning}
Online metric learning has been widely explored for instance evolution to learn large-scale streaming data, which is mainly composed of Mahalanobis distance-based and bilinear similarity-based methods. For the Mahalanobis distance-based methods, POLA \cite{ShalevShwartz2004} is the first attempt to learn the optimal metric in an online manner. Then several variants \cite{NIPS2008_3446, Davis2007IML1273496, 8617698} extend this idea by the fast similarity searching strategies, \emph{e.g.}, \cite{NIPS2009_3703} proposes a regularized online metric learning model with the provable regret bound. Besides, pairwise constraint~\cite{NIPS2009_3703} and triplet constraint \cite{NIPS2011_4392} are adopted to learn a discriminative metric function. Generally, triplet constraints perform better than pairwise constraints to learn a discriminative metric function \cite{NIPS2011_4392, Qian2015}.
For the bilinear similarity-based models, OASIS \cite{Chechik:2010:LSO:1756006.1756042} is developed to explore a similarity metric for recognition tasks, and SOML \cite{Gao2014SOMLSO} aims to learn a diagonal matrix for high dimensional cases with the similar setting as OASIS \cite{Chechik:2010:LSO:1756006.1756042}. \cite{6579606} presents an online multiple kernel similarity to tackle multi-modal tasks.

Unfortunately, these recently-proposed online metric learning methods cannot exploit the discriminative similarity relations for the strictly unaligned heterogeneous data in different evolution stages. 
To explore heterogeneous relationships among different data samples, \cite{7258342} focuses on learning a nonlinear metric to distinguish the foreground boundary and background for robust visual tracking. Duan \emph{et al.} \cite{7937828} design fine-grained localized distance metrics to learn hierarchical nonlinear transformations between heterogeneous samples. Ding \emph{et al.} \cite{10.1109/TCSVT.2018.2879626} introduce the fast low-rank learning mechanism and representation denoising strategy to explore a more robust metric learning framework.
Furthermore, \cite{7529190} proposes a multi-modal distance metric method for image ranking by incorporating both click and visual representations in distance metric learning. \cite{6774452} presents a multi-view stochastic learning model with high-order distance metric to explore modality-specific statistical information.
However, above-mentioned metric methods cannot be successfully applied to the challenging online scenarios, where the features are evolving due to the different senor lifespans (\emph{e.g.}, some features are vanished and some new features are augmented).

\subsection{Feature Evolution}
For the feature evolution, with the assumption that there exists samples from both vanished feature space and augmented feature space in an overlapping period, \cite{Hou:2017:LFE:3294771.3294906} develops an evolvable feature learning model by reconstructing the vanished features and exploiting it along with new emerging features for large-scale streaming data. \cite{Hou2018OnePassLW} proposes an one-pass incremental and decremental learning model for streaming data, which consists of a compressing stage and a expanding stage. Different from \cite{Hou:2017:LFE:3294771.3294906}, \cite{Hou2018OnePassLW} assumes that there are overlapping features instead of overlapping period. Similar to \cite{Hou2018OnePassLW}, \cite{Ye2018RectifyHM} focuses on learning the mapping function from two different feature spaces by using optimal transport technique. Furthermore, \cite{Zhang2015TowardsMT,7465766} intend to classify trapezoidal data stream with feature and instance increasing doubly. However, the new emerging samples often have overlapping features with the previously existing samples. \cite{8410016} develops an incremental feature learning model to tackle the emergence of new activity recognition sensors, which encourages the proposed model to well generalize the sudden emergence of incremental features.

Amongst the discussion above, there are no any feature evolution models highly related to our work except for OPID (OPIDe) \cite{Hou2018OnePassLW}. However, there are several key differences between \cite{Hou2018OnePassLW} and our EML model: 1) Our work is the first attempt to explore both instance and feature evolutions simultaneously via T-stage and I-stage in the metric learning field, when compared with \cite{Hou2018OnePassLW}. 2) Due to the strictly unaligned evolving features in the different stages, we utilize the smoothed Wasserstein distance to explore the distance relationships among the heterogeneous and complex data, rather than the Euclidean distance in \cite{Hou2018OnePassLW}. 3) Compared with \cite{Hou2018OnePassLW}, the low-rank regularizer for distance matrix could effectively learn a discriminative low-rank metric space, while neglecting non-informative knowledge for heterogeneous data in different feature evolution stages.

\section{Evolving Metric Learning (EML)} \label{sec:formulation}
This section first reviews online metric learning, and then detailedly introduces how to tackle both instance and feature evolutions via our proposed EML model.

\subsection{Revisit Online Metric Learning}
Metric learning focuses on exploring an optimal distance metric matrix, in the light of different measure functions, \emph{e.g.}, Mahalanobis distance function: $d_M(x_p, x_q) = \sqrt{(x_p-x_q)^{\top}M(x_p-x_q)}$, where $x_p\in\mathbb{R}^d$ and $x_q\in\mathbb{R}^d$ are the $p$-th and $q$-th samples, respectively. $M\in\mathbb{R}^{d\times d}$ is the symmetric positive semi-definite matrix, which can be formulated as $L^{\top}L$ \cite{NIPS2008_3446}, where $L\in\mathbb{R}^{r\times d}$ ($r$ denotes the rank of $M$) is the transformation matrix. The Mahalanobis distance function between $x_p$ and $x_q$ can be rewritten as $d_L(x_p, x_q) = \left\|L(x_p-x_q)\right\|_2^2$. Given an online constructed triplet $(x_p, x_q, x_k)$, $L$ could be updated in an online manner via the Passive-Aggressive algorithm \cite{Crammer:2006:OPA:1248547.1248566}, \emph{i.e.},
\begin{equation}\label{eq:M_formulation}
L_t = \arg\min\limits_{L} \frac{1}{2} \left\|L-L_{t-1} \right\|_F^2 + \frac{\gamma}{2} \ell_L(x_p, x_q, x_k),
\end{equation}
where $\ell_L(x_p, x_q, x_k)  = \big[1+d_L(x_p, x_q)-d_L(x_p, x_k) \big]_+$ is a hinge loss function. $[z]_+ = \mr{max}(0, z)$. $x_p$ and $x_q$ belong to the same class, and $x_p$ and $x_k$ belong to different classes. $\gamma\geq0$ is the regularization parameter.

However, most existing online metric learning models only focus on instance evolution with a fixed feature dimensionality, which cannot be utilized in the feature evolution scenario, \emph{i.e.}, streaming data with incremental and decremental features. 
Furthermore, they mainly aim to promote the discrimination of the learned distance matrix $L$ by minimizing the squared Mahalanobis distance from similar sample pairs. Especially, they assume that the feature descriptors of the sample pairs they focus on addressing are often aligned well in advance. Unfortunately, due to some unavoidable factors like non-linear lighting changes, heavily intensity noise and geometrical deformation, such assumption is heavily violated in the real-world tasks, especially for the feature evolution tasks. Therefore, the learned distance matrix $L$ in Eq.~\eqref{eq:M_formulation} is not applicable and discriminative to explore similarity relationships between the heterogeneous and complex samples, whose evolving feature descriptors are not strictly aligned in different evolution stages \cite{xu2018multi}.

\subsection{The Proposed EML Model}
This subsection first introduces how to integrate a smoothed Wasserstein distance into online metric formulation (\emph{i.e.}, Eq.~\eqref{eq:M_formulation}) to characterize the similarity relations of heterogeneous data with feature evolution in the different stages. 
Then the details about how to tackle feature evolution via Transforming stage (T-stage) and Inheriting stage (I-stage) in one-shot scenario are elaborated, followed by the extension of multi-shot scenario.

\subsubsection{Online Wasserstein Metric Learning} 
Wasserstein distance \cite{5703094} is an optimal transportation to transport all the earth from the source to target destination, while requiring the minimum amount of efforts. Formally, given two signatures $P = \{(x_{pi}, \mu_{pi})\}_{i=1}^m$ and $Q = \{(x_{qj}, \mu_{qj})\}_{j=1}^n$, the smoothed Wasserstein distance \cite{Cuturi:2013:SDL:2999792.2999868} between $P$ and $Q$ is: 
\begin{equation}\label{eq:smooth_W} 
\begin{aligned}
W_{\sigma}(P, Q)& = \min\limits_{F\in\mathbb{F}(P, Q)} \left<D(P, Q), F\right> - \sigma h(F),  \\
s.t.~\mathbb{F}(P, Q)&=\{F|F\textbf{1}_n=\mu_p, F^{\top}\textbf{1}_m=\mu_q, F\ge 0\},
\end{aligned}
\end{equation}
where $D(P, Q)=\{d_L(i, j)\}_{i,j=1}^{m,n}\in\mathbb{R}^{m\times n}$, and $d_L(i, j)$ denotes the cost of transporting one unit of earth from the source sample $x_{pi}$ to the target sample $x_{qj}$. $F = \{f(i, j)\}_{i,j=1}^{m,n}$ indicates the flow network matrix, and $f(i, j)$ represents the amount of earth that is transported from $x_{pi}$ to $x_{qj}$. $\mu_p = [\mu_{p1}, \cdots, \mu_{pm}]\in \mathbb{R}^m$ and $\mu_q = [\mu_{q1}, \cdots, \mu_{qn}] \in \mathbb{R}^n$ are normalized marginal probability mass vectors, and they satisfy $\sum_i \mu_{pi} =1$ and $\sum_j \mu_{qj} = 1$. $\sigma \geq0$ is a balance parameter, and $h(F)= -\left<F, \log(F)\right>$ is the strictly concave entropic function.

In Eq.~\eqref{eq:smooth_W}, the Mahalanobis distance is employed as ground distance to construct smoothed Wasserstein distance. Thus, each element $d_L(i, j)$ of $D(P, Q)$ in Eq.~\eqref{eq:smooth_W} represents the squared Mahalanobis distance between the source sample $x_{pi}$ of $P$ and the target sample $x_{qj}$ of $Q$, \emph{i.e.}, $d_L(i, j) = \left\|L(x_{pi}-x_{qj})\right\|_2^2$. Given the online constructed triplet $(P, Q, K)$ via \cite{pmlr-v51-rolet16}, where the samples of $P$ and $Q$ belong to the same class, and the samples of $P$ and $K$ belong to different classes. After substituting Mahalanobis distance in Eq.~\eqref{eq:M_formulation} with the smooth Wasserstein distance defined in Eq.~\eqref{eq:smooth_W}, online Wasserstein metric learning could be formulated as follows:
\begin{equation}\label{eq:W_formulation}
\begin{aligned}
\!\!\!\!\!\! \min\limits_{L, F} \mc{L}_L(P, Q, K)\! = \frac{1}{2} \left\|L\!-\!L_{t-1} \right\|_F^2+\!\frac{\gamma}{2}\ell_{L}(P,Q,K),
\end{aligned}
\end{equation}
where $\ell_L(P,Q,K)=[1 + W_{\sigma}(P, Q) - W_{\sigma}(P, K)]_+.$ When compared with the triplet $(x_p, x_q, x_k)$, each signature in $(P, Q, K)$ consists of several samples belonging to same class rather than only one sample.

\begin{figure}[t]
	\centering
	\includegraphics[scale=0.728]{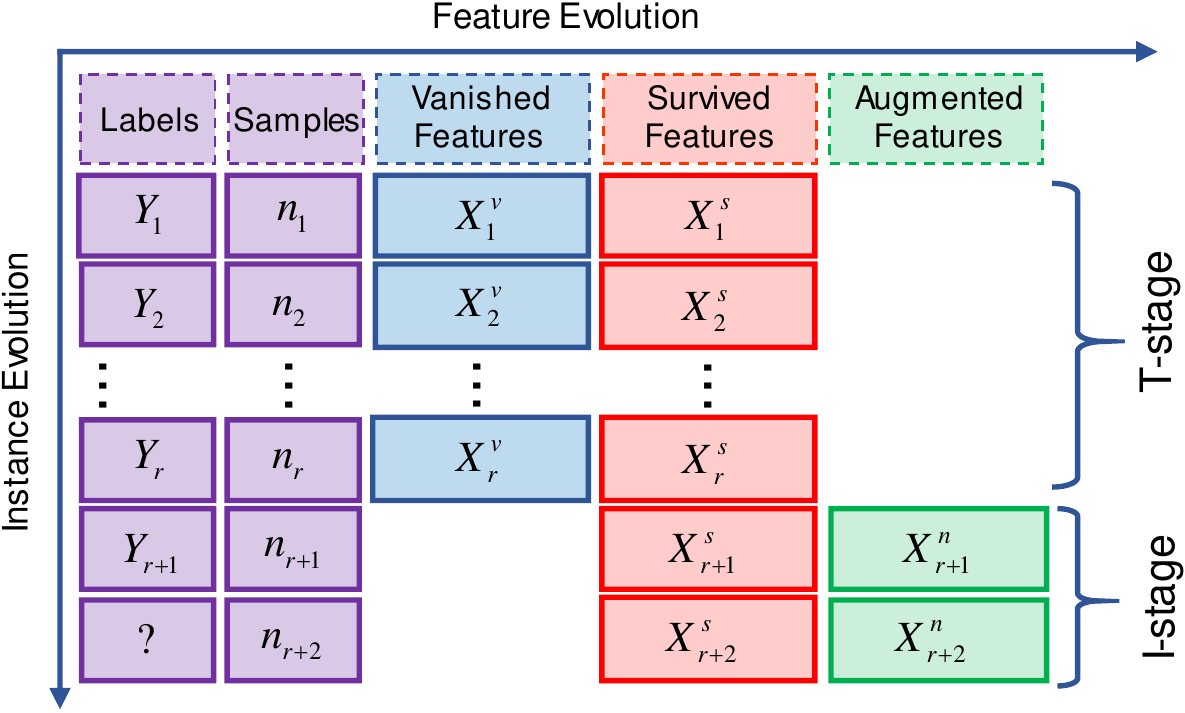} 
	\vspace{-15.0pt}
	\caption{The illustration of our EML model in one-shot scenario, which evolves instances and features simultaneously via T-stage and I-stage. Different colors denote different kinds of features, \emph{e.g.}, blue, red and green colors denote the vanished, survived and augmented features, respectively. The purple color indicates labels and the number of corresponding samples.}
	\label{fig:main_algorithm}
\end{figure}

\subsubsection{Transforming Stage (T-stage) $\&$ Inheriting Stage (I-stage)} 
In one-shot scenario where the features of streaming data would only be incremental and decremental by one time, two essential stages (\emph{i.e.}, T-stage and I-stage) of our proposed EML model for steaming data with feature evolution are elaborated below.

\textbf{I. Transforming Stage (T-stage):} As shown in Fig.~\ref{fig:main_algorithm}, suppose that $\{X_i, Y_i\}_{i=1}^r$ denotes the streaming data in the T-stage, where $X_i = [X_i^v, X_i^s]\in\mathbb{R}^{n_i \times (d_v + d_s)}$ and $Y_i\in\mathbb{R}^{n_i}$ denote the samples and labels in the $i$-th batch, respectively. $r$ is the total batches in T-stage and $n_i$ indicates the sample number in the $i$-th batch. Obviously, each instance of $X_i$ consists of vanished and survived features, and $d_v$ and $d_s$ indicate the corresponding dimensions of vanished features $X_i^v\in\mathbb{R}^{n_i\times d_v}$ and survived features $X_i^s\in\mathbb{R}^{n_i\times d_s}$.

If we directly combine both vanished and survived features to learn a unified metric function, it fails to be utilized in I-stage where some features are vanished and some other new features are augmented. We thus propose to extract important information from vanished features and forward it into survived features by exploring a common discriminative metric space. In other words, we aim to train a model using only survived features to characterize the effective information extracted from both vanished and survived features.

In the $i$-th batch of T-stage, inspired by \cite{pmlr-v51-rolet16}, the triplet $(P_i^s, Q_i^s, K_i^s)$ for survived features is constructed in an online manner, where the samples of $P_i^s\in\mathbb{R}^{n_p\times d_s}$ and $Q_i^s\in\mathbb{R}^{n_q\times d_s}$ belong to same class while the samples of $P_i^s$ and $K_i^s\in\mathbb{R}^{n_k\times d_s}$ belong to different classes. $n_p, n_q$ and $n_k$ are the numbers of samples in each signature. Likewise, we can construct the triplet $(P_i^a, Q_i^a, K_i^a)$ for all features (containing both vanished and survived features) in T-stage, where the samples of $P_i^a\in\mathbb{R}^{n_p\times (d_v+d_s)}$ and $Q_i^a\in\mathbb{R}^{n_q\times (d_v+d_s)}$ belong to same class while the samples of $P_i^a$ and $K_i^a\in\mathbb{R}^{n_k\times (d_v+d_s)}$ belong to different classes.

Let $L^s\in\mathbb{R}^{k\times d_s}$ and $L^a\in\mathbb{R}^{k\times (d_v+d_s)}$ denote the distance matrices trained on survived features and all features (containing both vanished and survived features) in T-stage. Since the dimensions of $L^s$ and $L^a$ are different, it is reasonable to add some essential consistency constraints on the optimal distance matrices $L^s$ and $L^a$ to extract important information from vanished features, and forward it into survived features. Generally, based on the smoothed Wasserstein metric learning in  Eq.~\eqref{eq:W_formulation}, the formulation of the $i$-th batch in the T-stage could be expressed as follows:
\begin{equation}\label{eq:objective_T-stage}
\begin{aligned}
& \!\!\!\!\!\!\!\!\!  \min\limits_{L^s, L^a, F} \mc{L}_{L^s}(P_i^s, Q_i^s, K_i^s) + \mc{L}_{L^a} (P_i^a, Q_i^a, K_i^a) +  \\
&\! \rho \mc{C}_{L^s,L^a}(P_i^s, Q_i^s, K_i^s; P_i^a, Q_i^a, K_i^a) + \lambda \mr{rank}(L^s, L^a),
\end{aligned} 
\end{equation}
where $\mc{L}_{L^s}(P_i^s, Q_i^s, K_i^s)$ and $\mc{L}_{L^a} (P_i^a, Q_i^a, K_i^a )$ denote the triplet losses of smoothed Wasserstein metric learning on survived features and all features (containing both vanished and survived features), respectively. $\mr{rank}(\cdot) = \mr{rank}(L^s) + \mr{rank}(L^a)$ denotes the regularization term, which learns the underlying low-rank property of heterogeneous samples. $\rho\geq0$ and $\lambda\geq 0$ are the balance parameters. $\mc{C}_{L^s,L^a}(\cdot;\cdot)$ in Eq.~\eqref{eq:objective_T-stage} is designed to enable the consistence constraint for $L^s$ and $L^a$, which aims to use only survived features to characterize the efficient information extracted from both vanished and survived features.

Specifically, $\mc{C}_{L^s,L^a}(\cdot;\cdot)$ constructs an essential triplet loss by incorporating smoothed Wasserstein metric learning on different feature spaces, \emph{i.e.}, survived features and all features (containing both vanished and survived features). We attempt to compute the smoothed Wasserstein distance between different heterogeneous distributions based on vanished features and all features. For example, $W_{\sigma}(P_i^a, Q_i^s) = \{d_L(u, v)\}_{u, v = 1}^{n_p, n_q}\in\mathbb{R}^{n_p\times n_q}$ denotes the smoothed Wasserstein distance between $P_i^a$ from all features and $Q_i^s$ from survived features, where $d_L(u, v) = \left\| L^a x_{pu}^a - L^s x_{qv}^s\right\|_2^2$ indicates the Mahalanobis distance between the $u$-th source sample $x_{pu}^a$ of $P_i^a$ and the $v$-th target sample $x_{qv}^s$ of $Q_i^s$. Likewise, $W_{\sigma}(P_i^a, K_i^s), W_{\sigma}(P_i^s, Q_i^a)$ and $W_{\sigma}(P_i^s, K_i^a)$ have similar definitions with $W_{\sigma}(P_i^a, Q_i^s)$. Formally, the consistence constraint $\mc{C}_{L^s,L^a}(\cdot;\cdot)$ is concretely expressed as follows:
\begin{equation}\label{eq:D_definition_constraint}
\begin{aligned}
\mc{C}_{L^s,L^a}(\cdot;\cdot) = &\big[W_{\sigma}(P_i^a, Q_i^s) - W_{\sigma}(P_i^a, K_i^s) + 1 \big]_+ \\ +&\big[W_{\sigma}(P_i^s, Q_i^a) - W_{\sigma}(P_i^s, K_i^a) + 1 \big]_+.
\end{aligned}
\end{equation}

\textbf{II. Inheriting Stage (I-stage):} Suppose that $\{X_{r+1}, Y_{r+1}\}$ denotes the data samples in the $r+1$-th batch of I-stage, where $X_{r+1} = [X_{r+1}^s, X_{r+1}^n]\in\mathbb{R}^{n_{r+1}\times (d_s + d_n)}$ indicates the samples and $Y_{r+1} \in\mathbb{R}^{n_{r+1}}$ is the corresponding labels, as shown in Fig.~\ref{fig:main_algorithm}.  $X_{r+1}^s$ and $X_{r+1}^n$ represent the survived features and new augmented features in the $r+1$-th batch. $d_n$ and $n_{r+1}$ are the dimension of the new augmented features and the number of samples. Thus, the goal of I-stage is to use $\{X_{r+1}, Y_{r+1}\}$ for training and make the prediction for the $r+2$-th batch data $X_{r+2}=[X_{r+2}^s, X_{r+2}^n]\in\mathbb{R}^{n_{r+2}\times (d_s + d_n)}$ whose number of samples is same as that of $X_{r+1}\in\mathbb{R}^{n_{r+1}\times (d_s + d_n)}$.

To classify the $r+2$-th batch data, we propose to inherit the metric performance of optimal distance matrix $L^s$ learned on survived features in T-stage, since a set of common survived features exist in both T-stage and I-stage. Although we could construct the triplets directly from the $r+1$-th batch for training, this trivial strategy has two significant shortcomings: 1) the trained metric model is difficult to be extended into multi-shot scenario; 2) the metric model learned only with the $r+1$-th batch data would have worse prediction performance due to the lack of full usage of data in T-stage.

To this end, we utilize a similar stacking strategy with \cite{Breiman1996,Zhou:2012:EMF:2381019}, where \cite{Breiman1996,Zhou:2012:EMF:2381019} focus on forming linear combinations of different predictors to train a unified classifier and achieve improved prediction accuracy. However, we propose to concatenate all feature descriptors as in stacking and train a unified predictor on the stacked features. It could better inherit the metric performance learned in T-stage. 
Concretely, let $Z_{r+1}^s = X_{r+1}^s(L^s)^{\top}\in\mathbb{R}^{n_{r+1}\times k}$ as the transformed discriminative metric space, which can be regarded as the new representation of $X_{r+1}^s$ for stacking. $X_{r+1}$ could then be represented as $Z_{r+1} = [Z_{r+1}^s, X_{r+1}^n]\in\mathbb{R}^{n_{r+1}\times(k+d_n)}$. Likewise, $X_{r+2}$ is characterized as $Z_{r+2}$. Furthermore, we learn an optimal distance matrix $L^z\in\mathbb{R}^{k\times(k+d_n)}$ on $Z_{r+1}$ with online constructed triplet $(P_{r+1}^z, Q_{r+1}^z, K_{r+1}^z)$, and evaluate the performance on $Z_{r+2}$, where the samples of $P_{r+1}^z$ and $Q_{r+1}^z$ belong to same class while the samples of $P_{r+1}^z$ and $K_{r+1}^z$ belong to different classes. Formally, at the $t$-th iterative step, the objective function of learning $L^z\in\mathbb{R}^{k\times(k+d_n)}$ in I-stage can be formulated as:
\begin{equation}\label{eq:objective_I-stage}
\begin{aligned}
&\!\!\!\!\!\! \min\limits_{L^z, F}\frac{1}{2} \left\|L^z - L_{t-1}^z \right\|_F^2 + \lambda \mr{rank}(L^z) \\ 
&+ \frac{\gamma}{2} \big[W_{\sigma}(P_{r+1}^z, Q_{r+1}^z) - W_{\sigma}(P_{r+1}^z, K_{r+1}^z) + 1 \big]_+,
\end{aligned}
\end{equation}
where $\gamma\geq 0$ and $\lambda\geq 0$ are the balance parameters. In our experiments, $\lambda$ and $\gamma$ in both Eq.~\eqref{eq:objective_T-stage} and Eq.~\eqref{eq:objective_I-stage} are set as the same value for simplification. $\mr{rank}(L^z)$ denotes the regularization term, which aims to explore the intrinsic low-rank structure of heterogeneous samples in I-stage.

\begin{figure}[t]
	\centering
	\includegraphics[scale=0.61] {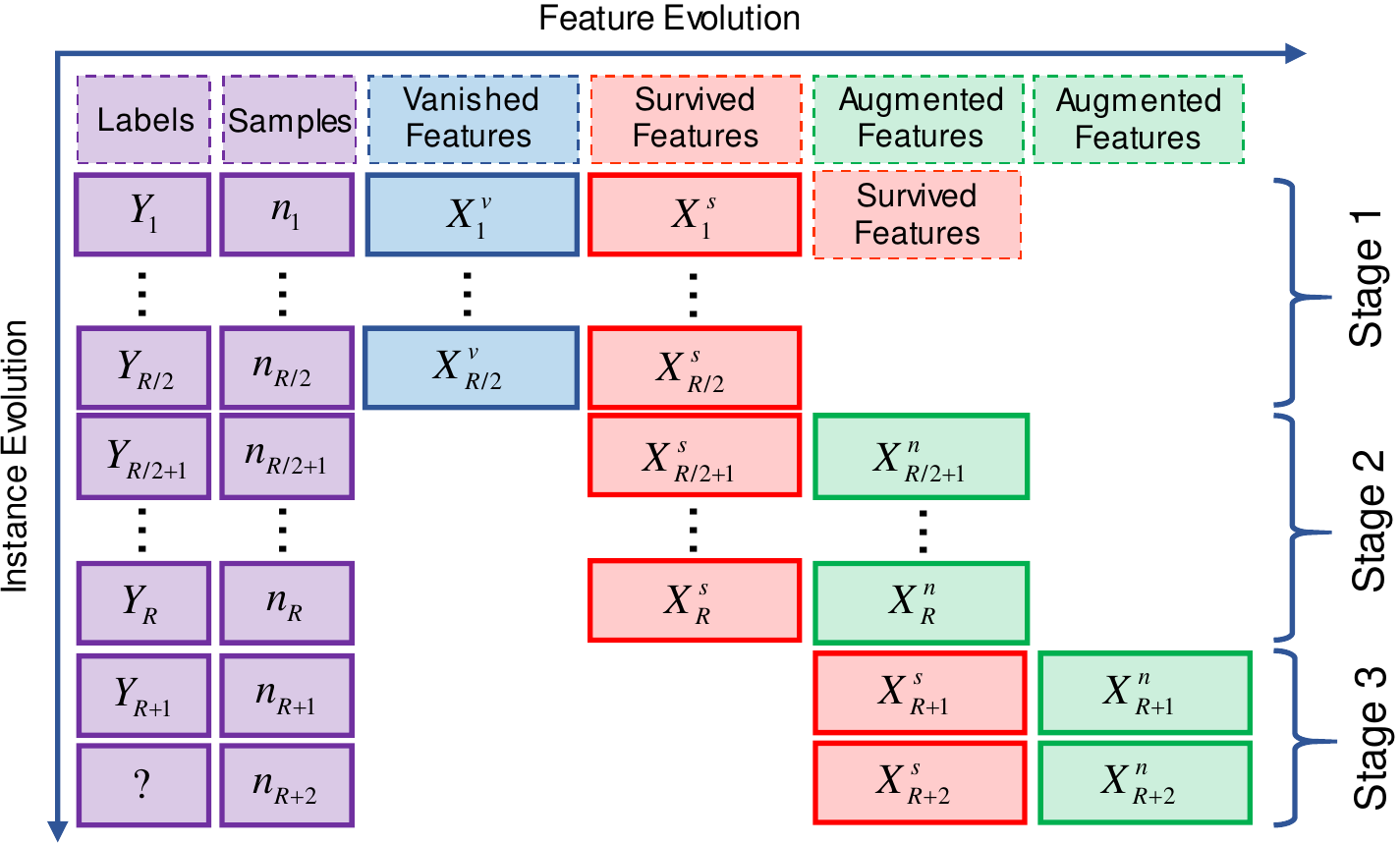} 
	\vspace{-15.0pt}
	\caption{The illustration of our EML model in multi-shot scenario when $M=2$, where Stage 1 and Stage 2 share the survived features, and Stage 2 and Stage 3 share the new augmented features. Specifically, our proposed model respectively regards Stage 1 and Stage 2 as T-stage and I-stage for the first feature evolution, and considers Stage 2 and Stage 3 as T-stage and I-stage for the second feature evolution.}
	\label{fig:multi-shot}
\end{figure}

\subsubsection{Multi-shot Scenario}  
Different from one-shot scenario, the features of streaming data in multi-shot scenario would be incremental and decremental $M$ times. This subsection extends our model from one-shot case into multi-shot scenario, and the illustration example of multi-shot scenario when $M=2$ is depicted in Fig.~\ref{fig:multi-shot}. Specifically, $\{X_i, Y_i\}_{i=1}^{R/2}$ denotes the streaming data in Stage 1, where $X_i = [X_i^v, X_i^s]\in\mathbb{R}^{n_i \times (d_v + d_s)}$ and $Y_i\in\mathbb{R}^{n_i}$ respectively represent the samples and labels in the $i$-th batch. $n_i$ indicates the sample number in the $i$-th batch, and $R/2$ denotes the total batches in Stage 1. When the streaming data $\{X_i, Y_i\}_{i=R/2+1}^{R}$ in Stage 2 arriving, it performs features evolution for the first time (\emph{i.e.}, some features are vanished and some new features are augmented), where $X_i = [X_i^s, X_i^n]\in\mathbb{R}^{n_i \times (d_s + d_n)}$. Moreover, in Stage 3, the streaming data $\{X_{R+1}, Y_{R+1}\}$ performs features evolution for the second time, and we predict the results of our proposed EML model on the $R+2$-th batch data $X_{R+2}$, where $X_{R+1} = [X_{R+1}^s, X_{R+1}^n]$ and $X_{R+2} = [X_{R+2}^s, X_{R+2}^n]$. Note that there are overlapped feature representations between any two adjacent stages. For example, as presented in Fig.~\ref{fig:multi-shot}, the survived features in Stage 1 are regarded as the vanished features in Stage 2, and the augmented feature in Stage 2 are considered as the survived features in Stage 3. Therefore, there are multiple Transforming stages (T-stage) and Inheriting stages (I-stage) in multi-shot scenario. To be specific, our proposed model first regards Stage 1 and Stage 2 as T-stage and I-stage for the first feature evolution. Then, it considers Stage 2 and Stage 3 as T-stage and I-stage for the second feature evolution. Generally, in multi-shot scenario, we have two essential learning tasks:
\begin{itemize}
	\setlength{\itemsep}{1pt}
	\setlength{\parsep}{0pt}
	\setlength{\parskip}{0pt}
	\item Task I: Similar to the prediction task in one-shot case, we aim to classify testing data $X_{R+2}$ in Stage 3 by training our proposed model on previous $R+1$ batch streaming data $\{X_i, Y_i\}_{i=1}^{R+1}$.  
	\item Task II: Different from the prediction task in one-shot scenario, we attempt to make predictions for all stages (\emph{i.e.}, Stage 1, Stage 2 and Stage 3 when $M=2$) by training our proposed model on the streaming data $\{X_i, Y_i\}_{i=1}^{R+1}$ in all stages. 
\end{itemize}

\section{Model Optimization} 
This section presents an alternating optimization strategy to update our proposed EML model amongst two stages, \emph{i.e.}, T-stage and I-stage, followed by the computational complexity analysis of our model. The whole optimization strategy of our proposed EML model is introduced in \textbf{Algorithm 1}.

Note that the low-rank minimization in Eq.~\eqref{eq:objective_T-stage} and Eq.~\eqref{eq:objective_I-stage} is a well-known NP hard problem. Take $L^z$ as an example, $\mr{rank}(L^z)$ in Eq.~\eqref{eq:objective_I-stage} can be effectively surrogated by trace norm $\left\| L^z \right\|_{*}$. Different from traditional Singular Value Thresholding (SVT) \cite{doi:10.1137/080738970}, we employ a regularization term to guarantee the low-rank property, \emph{i.e.}, $\left\| L^z \right\|_{*} = \mr{tr}\big(({L^z}^{\top}L^z)^{1/2}\big) = \mr{tr}\big({L^z}^{\top}(L^z{L^z}^{\top})^{-1/2} L^z \big)$. As a result, $\mr{rank}(L^z)$ in Eq.~\eqref{eq:objective_I-stage} could be formulated as $\mr{tr}({L^z}^{\top}H^z{L^z})$, where $H^z = (L^z {L^z}^{\top})^{-1/2}$. Likewise, the low rank optimization of $L^a$ and $L^s$ shares the same strategy with $L^z$. $\mr{rank}(L^a)$ and $\mr{rank}(L^s)$ are respectively surrogated by $\mr{tr}(L^aH^a{L^a}^{\top})$ and $\mr{tr}(L^sH^s{L^s}^{\top})$, where $H^a = (L^a {L^a}^{\top})^{-1/2}$ and $H^s = (L^s {L^s}^{\top})^{-1/2}$.

\renewcommand{\algorithmicrequire}{\textbf{Input:}}		
\renewcommand{\algorithmicensure}{\textbf{Output:}}		
\begin{algorithm}[t]	
	\caption{\small The Optimization of Our Proposed EML Model} 
	\begin{algorithmic}[1]
		\REQUIRE The data $\{X_i, Y_i\}_{i=1}^{r+1}$, the parameters $\gamma, \lambda, \rho$;
		\ENSURE $L^s$ and $L^z$;   \\ 
		\STATE \textbf{Initialize}: $L^s, L^a, L^z, F$;
		\STATE \begin{elaboration} \textbf{Transforming stage (T-stage):} \end{elaboration} 
		\FOR {$i=1,\ldots,r$} 
		\STATE Calculate the smoothed Wasserstein distance for data $X_{i}$, and construct the triplets for training; 
		\REPEAT 
		\STATE Solve $F$ when fixing $L^a$ and $L^s$; 
		\STATE Update $L^a$ via Eq.~\eqref{eq:solution_T_L_all}; 
		\STATE Update $L^s$ via Eq.~\eqref{eq:solution_T_L_s}; 
		\STATE Update $H^a$ and $H^s$ via $H^a = (L^a{L^a}^{\top})^{-1/2}$ and $H^s = (L^s{L^s}^{\top})^{-1/2}$;
		\UNTIL{Converge}
		\ENDFOR	
		
		\STATE \begin{elaboration} \textbf{Inheriting stage (I-stage):} \end{elaboration}
		\STATE Transform $X_{r+1}$ as $Z_{r+1}$ to calculate  smoothed Wasserstein distance, and construct the training triplets;
		\REPEAT
		\STATE Solve the distance flow-network $F$ when fixing $L^z$;
		\STATE Update $L^z$ via Eq.~\eqref{eq:solution_I_L_all};
		\STATE Update $H^z$ via $H^z = (L^z{L^z}^{\top})^{-1/2}$;
		\UNTIL{Converge}
	\end{algorithmic}					
\end{algorithm}

\subsection{Optimizing T-stage via an Alternating Strategy}

\subsubsection{Updating $L^a$ by fixing $\{L^s, H^a, F\}$} When fixing the variables $L^s, H^a$ and $F$, the optimization problem in Eq.~\eqref{eq:objective_T-stage} for solving variable $L^a$ can be concretely expressed as:
\begin{equation}\label{eq:optimize_T_L_all}
\begin{aligned}
& L_t^a = \arg\min\limits_{L^a} \frac{1}{2} \left\|L^a-L_{t-1}^a \right\|_F^2 +  \lambda \mr{tr}(L^aH^a{L^a}^{\top})  \\ 
&\; + \frac{\gamma}{2} \big[\mr{tr} \big(D(P_i^a, Q_i^a)F \big) - \mr{tr} \big(D(P_i^a, K_i^a)F\big) + 1 \big]_+   \\
& \; + \frac{\rho}{2} \big[\mr{tr} \big(D(P_i^a, Q_i^s)F\big) - \mr{tr} \big(D(P_i^a, K_i^s)F\big) + 1 \big]_+   \\
& \;+ \frac{\rho}{2} \big[\mr{tr} \big(D(P_i^s, Q_i^a)F\big) - \mr{tr} \big(D(P_i^s, K_i^a)F\big) + 1 \big]_+. \\
\end{aligned}
\end{equation}
The optimal solution of $L_t^a$ could be relaxedly   
achieved by nulling the gradient of Eq.~\eqref{eq:optimize_T_L_all}:
\begin{equation}\label{eq:solution_T_L_all}
\begin{aligned}
\!\!L_t^a = \big(L_{t-1}^a \! - \! \rho L_t^s(G_3 +G_4)\big)\big(\mr{I} + \lambda H^a + \gamma G_1 + \rho G_2\big)^{-1},
\end{aligned}
\end{equation}
where $ G_1 = {Q_i^a}^{\top} \mr{diag}(\textbf{1}^{\top}F){Q_i^a} - {K_i^a}^{\top} \mr{diag}(\textbf{1}^{\top}F){K_i^a} -
{P_i^a}^{\top} F{Q_i^a}- {Q_i^a}^{\top} F^{\top}{P_i^a} + {P_i^a}^{\top} F{K_i^a} +  {K_i^a}^{\top} F^{\top}{P_i^a}, G_2 = {P_i^a}^{\top}  \mr{diag}(F\textbf{1}){P_i^a} - {K_i^a}^{\top}  \mr{diag}(\textbf{1}^{\top}F){K_i^a}, G_3 = {K_i^s}^{\top} F{P_i^a} -  {Q_i^s}^{\top} F^{\top}{P_i^a}, G_4 = {P_i^s}^{\top} F{K_i^a} - {K_i^s}^{\top} F^{\top}{P_i^a}$.

\subsubsection{Updating $L^s$ by fixing $\{L^a, H^s, F\}$} With the obtained distance matrix $L^a$ and flow matrix $F$, the optimization problem for variable $L^s$ in Eq.~\eqref{eq:objective_T-stage} could be formulated as:
\begin{equation}\label{eq:optimize_T_L_s}
\begin{aligned}
& L_t^s = \arg\min\limits_{L^s} \frac{1}{2} \left\|L^s-L_{t-1}^s \right\|_F^2 + \lambda \mr{tr}(L^sH^s{L^s}^{\top}) \\
&  +\frac{\gamma}{2} \big[\mr{tr} \big(D(P_i^s, Q_i^s)F \big) - \mr{tr} \big(D(P_i^s, K_i^s)F\big) + 1 \big]_+   \\
& +\frac{\rho}{2} \big[\mr{tr} \big(D(P_i^a, Q_i^s)F\big) - \mr{tr} \big(D(P_i^a, K_i^s)F\big) + 1 \big]_+  \\
& +\frac{\rho}{2} \big[\mr{tr} \big(D(P_i^s, Q_i^a)F\big) - \mr{tr} \big(D(P_i^s, K_i^a)F\big) + 1 \big]_+ . \\  
\end{aligned}
\end{equation}
Concretely, the updating operator for $L_t^s$ could be given as:
\begin{equation}\label{eq:solution_T_L_s}
\begin{aligned}
\!\!L_t^s = \big(L_{t-1}^s- \rho L_t^a(G_6 +G_8)\big)\big(\mr{I} + \lambda H^s + \gamma G_5 + \rho G_7\big)^{-1},
\end{aligned} 
\end{equation} 
where $G_5 = {Q_i^s}^{\top} \mr{diag}(\textbf{1}^{\top}F){Q_i^s} - {K_i^s}^{\top} \mr{diag}(\textbf{1}^{\top}F){K_i^s} + {P_i^s}^{\top} F{K_i^s} + {K_i^s}^{\top} F^{\top}{P_i^s} - {P_i^s}^{\top} F{Q_i^s} - {Q_i^s}^{\top} F^{\top}P_i^s, G_6 = {P_i^a}^{\top} F{K_i^s} - {P_i^a}^{\top} F{Q_i^s}, G_7 = {Q_i^s}^{\top} \mr{diag}(F\textbf{1}){Q_i^s} - {K_i^s}^{\top}   \mr{diag}(\textbf{1}^{\top}F){K_i^s}, G_8 = {K_i^a}^{\top} F^{\top}{P_i^s} - {Q_i^a}^{\top} F^{\top}{P_i^s}$. 

\subsubsection{Updating $F$ by fixing $\{L^a, L^s\}$} When the distance matrices $L^a$ and $L^s$ are fixed, we split Eq.~\eqref{eq:objective_T-stage} into some independent traditional smoothed Wasserstein distance subproblems, which could be solved by \cite{pmlr-v51-rolet16}. We omit the detailed process of solving smoothed Wasserstein distance subproblems for simplification.

\subsection{Optimizing I-stage via an Alternating Strategy}

\subsubsection{Updating $L^z$ by fixing $\{H^z, F\}$} Given the fixed $H^z$ and $F$, the formulation for $L^z$ in Eq.~\eqref{eq:objective_I-stage} is rewritten as: 
\begin{equation}\label{eq:optimize_I_L_all}
\begin{aligned}
&L_t^z = \arg\min\limits_{L^z} \frac{1}{2}\left\|L^z - L_{t-1}^z \right\|_F^2 + \lambda \mr{tr}(L^zH^z{L^z}^{\top}) +  \\
& \frac{\gamma}{2} \big[ \mr{tr}\big(D(P_{r+1}^z, Q_{r+1}^z)F\big) - \mr{tr}\big(D(P_{r+1}^z, K_{r+1}^z)F\big) + 1 \big]_+.
\end{aligned}
\end{equation}
By nulling the gradient of Eq.~\eqref{eq:optimize_I_L_all}, the optimization solution of Eq.~\eqref{eq:optimize_I_L_all} for $L^z$ could be given as:
\begin{equation}\label{eq:solution_I_L_all}
L_t^z = L_{t-1}^z(\mr{I} + \lambda H^z + \gamma G_9)^{-1},
\end{equation}
where $G_9 = Q_{r+1}^{z^{\top}} \mr{diag}(\textbf{1}^{\top}F)Q_{r+1}^{z}+ P_{r+1}^{z^{\top}}FK_{r+1}^z - K_{r+1}^{z^{\top}}\mr{diag}(\textbf{1}^{\top}F)K_{r+1}^z
+ K_{r+1}^{z^{\top}}F^{\top}P_{r+1}^z -P_{r+1}^{z^{\top}}FQ_{r+1}^{z} - Q_{r+1}^{z^{\top}}F^{\top}P_{r+1}^{z}$.

\subsubsection{Updating $F$ by fixing $L^z$} The optimization procedure of variable $F$ in I-stage is same as that in T-stage: with the fixed $L^z$, the formulation Eq.~\eqref{eq:objective_I-stage} is split into some independent traditional smoothed Wasserstein distance subproblems, and we solve the variable $F$ via \cite{pmlr-v51-rolet16}.

\begin{figure}[t]
	\centering
	\includegraphics[width =252pt,height =105pt] {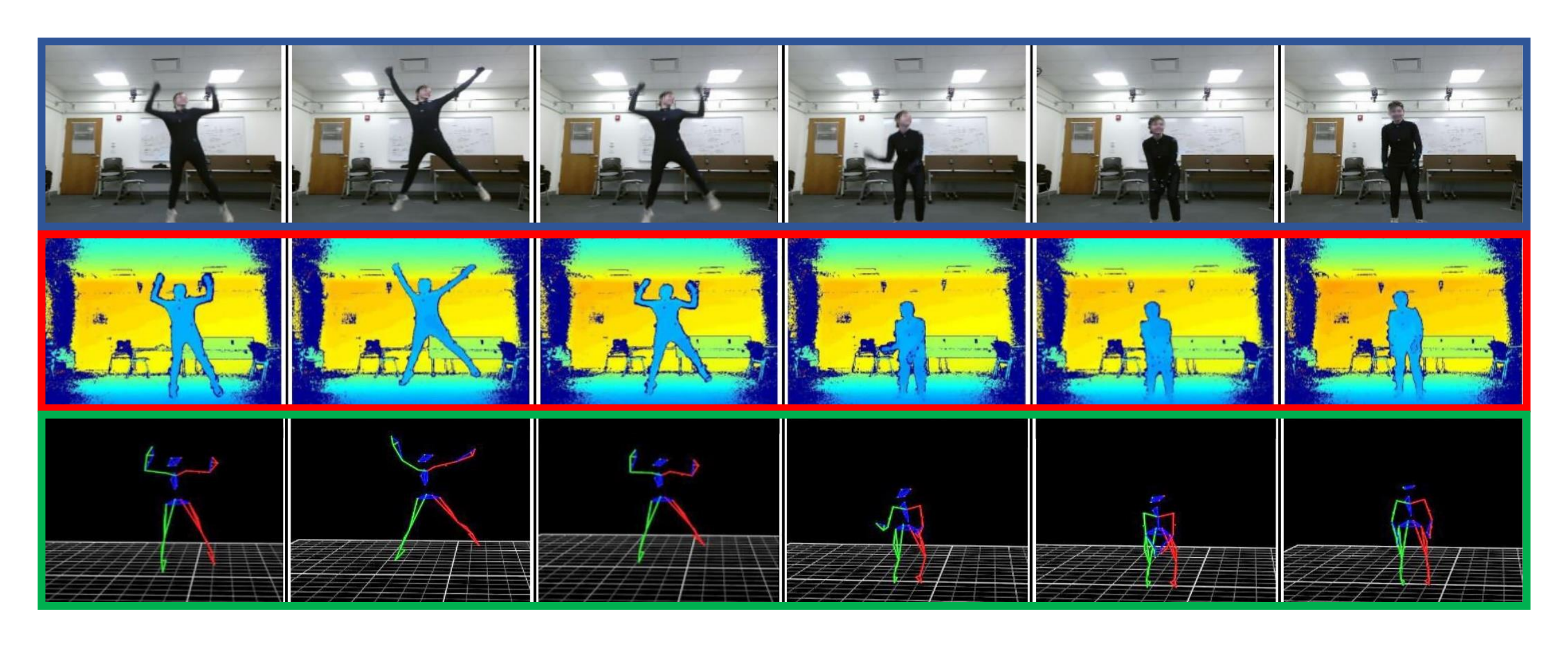}
	\vspace{-20pt}
	\caption{The examples of human motions in EV-Action dataset, where the first, second and third rows denote the samples collected from RGB camera, Kinect sensor and motion capture sensor, respectively.}
	\label{fig:examples_human_motion}
\end{figure}

\begin{table}[t]
	\centering
	\setlength{\tabcolsep}{1.25mm}
	\caption{The experimental settings in one-shot scenario.}
	\scalebox{0.93}{
		\begin{tabular}{|c|c|c|c|c|c|c|}
			\hline
			Datasets & c & $\sum_{i=1}^{r} n_i$ & $n_i$ & $d_v$ & $d_s$ & $d_n$  \\
			\hline
			EV-Action & 20 & 4200 & $500, \,600, \,700$ & 1024 & 1024 & 75 \\
			Mnist0vs5 & 2 & 3200 & $80, \,160,\,320$ & 114 & 228 & 113 \\
			Mnist0vs3vs5 & 3 & 4800 &$ 120,\,240,\,480$ & 123 & 245 & 121 \\			
			Splice & 2 & 2240 & $80,\,160,\,320$ & 10 & 40 & 10 \\
			Gisette & 2 & 6000 & $100,\,200,\,300$ & 1239 & 2478 & 1238 \\
			USPS0vs5 & 2 & 960 & $120,\,160,\,240$ & 64 & 128 & 64 \\
			USPS0vs3vs5 & 3 & 1440 & $180,\,240,\,300$ & 64 & 128 & 64 \\
			Satimage & 3 & 1080 & $60,\,90,\,120$ & 10 & 18 & 8 \\
			ImageNet & 1000 & 1200000 & $10000,\,12000,\,14000$ & 512 & 1024 & 512 \\
			PAMAP2  & 18 & 7200 & $600,\,700,\,800$ & 81 & 162 & 81 \\
			\hline	
		\end{tabular}
	}
	\label{table:settings_one_shot}
\end{table}

\begin{table*}[t]
	\centering
	\setlength{\tabcolsep}{1.45mm}
	\caption{Comparisons between our model and state-of-the-art methods in terms of accuracy (\%) on ten datasets: mean and standard errors averaged over fifty random runs in one-shot scenario. Models with the best performance are bolded.}
	\scalebox{0.938}{
		\begin{tabular}{|c|c|cccccc|ccc|c|}
			\hline
			Dataset & $n_i$ & Pegasos \cite{Shalev-Shwartz2011} & OPMV \cite{Zhu2015OnePassML} & TCA \cite{5640675} & BDML \cite{Xu:2018:BDM:3327144.3327333} & OPML \cite{LI2018302} & CDML \cite{Chen2019CurvilinearDM} & OPIDe \cite{Hou2018OnePassLW} & OPID \cite{Hou2018OnePassLW} & FIRF \cite{8410016} & Ours\\
			\hline
			
			& 500  & 57.38$\pm$1.51 & 56.37$\pm$1.91 & 53.88$\pm$2.04
			& 56.42$\pm$0.71
			& 54.10$\pm$1.71 & 55.08$\pm$0.83 & 57.84$\pm$1.06 & 57.57$\pm$1.08 & 57.13$\pm$0.84 & \textbf{58.87$\pm$0.68} \\
			EV-Action & 600 & 57.46$\pm$1.60 & 56.94$\pm$1.82 & 54.61$\pm$1.73
			& 56.81$\pm$0.65
			& 55.37$\pm$1.64 & 55.92$\pm$1.03 & 57.22$\pm$0.95 & 56.71$\pm$1.40 & 56.92$\pm$1.25 & \textbf{58.65$\pm$0.84} \\
			& 700 & 57.22$\pm$1.34 & 56.68$\pm$1.87  & 54.37$\pm$1.69
			& 56.63$\pm$0.77
			& 55.82$\pm$1.62 & 56.22$\pm$0.71 & 57.09$\pm$1.13 & 56.85$\pm$1.27 & 57.23$\pm$1.16 & \textbf{58.32$\pm$0.82} \\		
			
			\hline	
			Mnist    & 80  & 97.74$\pm$0.73 & 97.39$\pm$0.92 & 96.53$\pm$1.75
			& 97.00$\pm$1.66
			& 96.45$\pm$1.72 & 96.75$\pm$1.32 & 98.68$\pm$0.88 & 98.88$\pm$0.99 & 98.14$\pm$0.87 & \textbf{99.85$\pm$0.91}  \\
			0vs5     & 160 & 98.11$\pm$1.03 & 95.82$\pm$1.84 & 93.08$\pm$2.94
			& 98.25$\pm$0.80
			& 96.83$\pm$1.38 & 97.04$\pm$0.58 & 97.94$\pm$0.97 & 98.75$\pm$0.90 & 96.79$\pm$1.52 & \textbf{99.78$\pm$0.57} \\
			& 320 & 97.68$\pm$0.79 & 96.47$\pm$1.79 & 92.43$\pm$3.82
			& 98.24$\pm$0.75
			& 96.98$\pm$1.03 & 97.16$\pm$0.85 & 97.38$\pm$0.58 & 97.21$\pm$0.66 & 96.83$\pm$1.37 & \textbf{99.27$\pm$0.37}\\
			\hline
			Mnist    & 120 & 91.47$\pm$3.92 & 95.87$\pm$1.82 & 91.26$\pm$3.87
			& 92.23$\pm$2.86
			& 92.42$\pm$2.22 & 92.66$\pm$1.49 & 94.58$\pm$1.78 & 94.97$\pm$1.30 & 95.03$\pm$0.83 & \textbf{96.91$\pm$1.38} \\
			0vs3vs5  & 240 & 89.95$\pm$3.08 & 93.96$\pm$1.18 & 90.85$\pm$1.74
			& 92.87$\pm$1.40
			& 91.99$\pm$1.64 & 92.47$\pm$1.31 & 93.45$\pm$1.41 & 93.48$\pm$1.35 & 94.24$\pm$1.13 & \textbf{95.37$\pm$0.92} \\
			& 480 & 90.12$\pm$1.93 & 93.28$\pm$1.69 & 91.14$\pm$3.95
			& 93.21$\pm$1.06
			& 92.74$\pm$1.17 & 93.04$\pm$0.96 & 93.30$\pm$0.86 & 93.37$\pm$0.79 & 93.85$\pm$0.95 & \textbf{95.54$\pm$0.87}\\
			\hline			
			& 80  & 79.65$\pm$4.13 & 80.13$\pm$3.86 & 76.93$\pm$4.52
			& 65.65$\pm$5.53
			& 69.60$\pm$4.38 & 68.85$\pm$2.27 & 81.22$\pm$3.73 & 80.50$\pm$3.53 & 79.83$\pm$2.55 & \textbf{82.65$\pm$3.32} \\
			Splice   & 160 & 82.25$\pm$3.26 & 81.95$\pm$2.84 & 80.93$\pm$3.47
			& 71.55$\pm$4.07
			& 78.21$\pm$2.53 & 75.85$\pm$2.65 & 84.00$\pm$2.03 & 83.91$\pm$2.05 & 82.06$\pm$1.91 & \textbf{85.25$\pm$2.06} \\
			& 320 & 82.32$\pm$3.18 & 78.72$\pm$4.37 & 81.53$\pm$3.38
			& 72.16$\pm$3.40
			& 80.86$\pm$2.01 & 78.93$\pm$1.17 & 85.55$\pm$1.32 & 85.94$\pm$1.38 & 83.69$\pm$1.73 & \textbf{87.03$\pm$1.52} \\
			\hline
			& 100 & 97.53$\pm$1.33 & 95.27$\pm$2.85 & 94.11$\pm$3.35
			& 90.25$\pm$3.13
			& 94.17$\pm$3.02 & 93.71$\pm$2.39 & 97.14$\pm$1.28 & \textbf{97.56$\pm$1.26} & 94.21$\pm$0.96 & 97.29$\pm$1.25 \\
			Gisette  & 200 & 95.14$\pm$2.97 & 94.05$\pm$3.36 & 93.03$\pm$3.16
			& 91.50$\pm$1.25
			& 93.61$\pm$3.19 & 92.68$\pm$1.72 & 95.59$\pm$0.95 & 95.39$\pm$1.06 & 93.76$\pm$0.79 & \textbf{96.82$\pm$0.91} \\
			& 300 & 96.84$\pm$1.35 & 93.71$\pm$3.11 & 94.37$\pm$3.72
			& 93.83$\pm$2.12
			& 93.77$\pm$2.96 & 93.24$\pm$1.56 & 96.36$\pm$0.69 & 95.33$\pm$0.93 & 94.18$\pm$1.06 & \textbf{97.89$\pm$0.43} \\
			\hline
			USPS     & 120 & \textbf{98.52$\pm$1.67} & 95.27$\pm$2.67
			& 96.42$\pm$1.81 & 95.90$\pm$1.65
			& 93.72$\pm$2.32 & 94.74$\pm$2.46 & 96.17$\pm$1.44 & 96.51$\pm$1.25 & 95.85$\pm$1.33
			& 97.23$\pm$1.64 \\
			0vs5     & 160 & 97.84$\pm$0.82 & 95.65$\pm$1.72 & 95.46$\pm$2.13
			& 96.38$\pm$1.23
			& 93.04$\pm$4.05 & 95.21$\pm$1.57 & 96.78$\pm$1.31 & 96.93$\pm$1.00 & 95.75$\pm$1.12
			& \textbf{98.91$\pm$0.67} \\
			& 240 & 97.93$\pm$0.72 & 96.17$\pm$1.28 & 95.85$\pm$2.07
			& 96.78$\pm$1.18
			& 93.62$\pm$3.01 & 95.62$\pm$1.83 & 94.93$\pm$1.28 & 95.06$\pm$1.10 & 93.72$\pm$0.93 & \textbf{98.94$\pm$0.70} \\
			\hline
			USPS     & 180 & 94.68$\pm$1.20 & 92.46$\pm$1.07 & 93.88$\pm$1.37
			& 90.62$\pm$2.48
			& 92.06$\pm$1.64 & 91.85$\pm$1.62 & 94.47$\pm$1.77 & 94.13$\pm$1.92 & 94.63$\pm$1.45
			& \textbf{95.73$\pm$0.88} \\
			0vs3vs5  & 240 & 94.39$\pm$1.09 & 91.69$\pm$2.31 & 92.94$\pm$1.58
			& 91.48$\pm$1.68
			& 91.23$\pm$1.73 & 91.73$\pm$1.24 & 92.08$\pm$1.93 & 92.50$\pm$1.66 & 93.36$\pm$2.07
			& \textbf{95.52$\pm$1.26} \\
			& 300 & \textbf{95.47$\pm$0.94} & 92.25$\pm$1.60 & 93.26$\pm$1.44
			& 92.13$\pm$1.09
			& 91.60$\pm$1.71 & 92.07$\pm$1.36 & 92.95$\pm$1.12 & 92.67$\pm$1.46 & 93.18$\pm$1.54
			& 94.05$\pm$1.46 \\
			\hline
			& 60  & 94.25$\pm$2.56 & 96.48$\pm$1.47 & 97.25$\pm$1.08
			& 97.14$\pm$1.59
			& 97.47$\pm$1.59 & 97.39$\pm$1.46 & 98.17$\pm$2.19 & 97.60$\pm$2.31 & 97.92$\pm$2.05 & \textbf{99.20$\pm$0.91} \\
			Satimage & 90  & 96.49$\pm$1.49 & 96.83$\pm$1.18 & 96.52$\pm$1.32
			& 97.62$\pm$1.52
			& 97.69$\pm$1.16 & 97.84$\pm$1.31 & 98.58$\pm$1.12 & 97.29$\pm$2.08 & 98.16$\pm$1.85 & \textbf{99.71$\pm$1.06} \\
			& 120 & 98.03$\pm$1.13 & 97.38$\pm$1.94 & 97.12$\pm$1.87
			& 97.12$\pm$1.48
			& 97.15$\pm$1.49 & 97.22$\pm$1.63 & 98.45$\pm$1.14 & 96.85$\pm$1.94 & 97.24$\pm$1.36 & \textbf{99.52$\pm$1.07} \\
			\hline
			         & 10000 & 55.28$\pm$1.83 & 51.03$\pm$2.58 & 50.44$\pm$3.15 & 52.49$\pm$3.14 & 52.74$\pm$2.54 & 52.15$\pm$2.71 & 55.63$\pm$1.22 & 55.70$\pm$2.03 & 53.94$\pm$2.05 & \textbf{56.47$\pm$1.57} \\
			ImageNet & 12000 & 56.37$\pm$1.75 & 50.24$\pm$2.39 & 50.83$\pm$2.96 & 52.68$\pm$2.33 & 52.94$\pm$1.87 & 52.06$\pm$2.64 & 55.94$\pm$1.83 & 56.31$\pm$2.33 & 54.82$\pm$1.77 & \textbf{57.83$\pm$1.93} \\
			         & 14000 & 58.04$\pm$2.38 & 51.61$\pm$3.52 & 50.62$\pm$2.74 & 53.02$\pm$3.14 & 53.73$\pm$2.19 & 52.64$\pm$2.37 & 56.85$\pm$1.52 & 57.06$\pm$1.84 & 54.79$\pm$2.29 & \textbf{59.17$\pm$1.84} \\
			\hline
			& 600 & 91.64$\pm$1.08 & 89.85$\pm$1.33 & 85.73$\pm$1.84 & 87.67$\pm$1.74 & 86.23$\pm$1.81 & 87.92$\pm$2.16 & 92.17$\pm$0.93 & 92.64$\pm$1.05 & 93.56$\pm$0.84 & \textbf{95.27$\pm$0.71} \\
			PAMAP2 & 700 & 91.85$\pm$1.15 & 90.14$\pm$1.29 & 86.04$\pm$2.03 & 88.06$\pm$2.20 & 87.94$\pm$1.57 & 88.73$\pm$1.91 & 91.85$\pm$1.28 & 92.39$\pm$1.04 & 93.28$\pm$1.13 & \textbf{95.46$\pm$0.85} \\
			& 800 & 91.57$\pm$0.89 & 90.25$\pm$1.56 & 85.49$\pm$2.75 & 88.75$\pm$2.06 & 88.13$\pm$1.90 & 89.37$\pm$1.68 & 93.05$\pm$0.88 & 93.62$\pm$0.93 & 93.84$\pm$0.79 & \textbf{95.66$\pm$0.94} \\
			
			\hline	
		\end{tabular}
	}
	\label{table:results_one_shot}
\end{table*}

\subsection{Computational Complexity Analysis}
The main computational cost in our EML model involves the updating operations in both T-stage and I-stage. Specifically, in the T-stage, the computational costs of updating $L^s$ and $L^a$ are $O\big(kd_s + k(d_v+d_s)^2d_s + d_s^3\big)$ and $O\big(k(d_s+d_v) + kd_s^2(d_v+d_s) + (d_v+d_s)^3\big)$, respectively. For the I-stage, solving the variable $L_z$ in Eq.~\eqref{eq:objective_I-stage} takes $O\big((k + d_a)^3\big)$. Besides, the computational cost of solving $F$ in both T-stage and I-stage is $O(n_p^2n_q^2 + n_p^2n_k^2 + n_q^2n_k^2)$, where $n_p,n_q,n_k \ll n_i$. When compared with the feature dimension and sample number, the value of $k$ is often small, and thus our proposed model is efficient to optimize in an online manner.

\section{Experiments}
This section first presents detailed experimental configurations and competing methods. Then the experimental performance along with some analyses about our EML model in both one-shot and multi-shot cases are provided.

\begin{table}[t]
	\centering
	\setlength{\tabcolsep}{3.5mm}
	\caption{Ablation study of our EML model in one-shot scenario.}
	\scalebox{0.775}{
		\begin{tabular}{|c|c|ccc|c|}
			\hline
			Dataset & $n_i$ & Ours-woT & Ours-woI & Ours-woW & Ours \\
			\hline	
			
			& 500  & 56.68$\pm$1.74 & 54.36$\pm$1.61 & 57.93$\pm$0.85 & \textbf{58.33$\pm$0.76} \\ 
			EV-Action  & 600 & 56.23$\pm$1.81 & 55.70$\pm$1.49 & 57.70$\pm$1.04
			& \textbf{57.94$\pm$0.88} \\ 
			& 700 & 57.02$\pm$1.56 & 55.93$\pm$1.76 & 57.83$\pm$0.92
			& \textbf{58.12$\pm$0.86} \\ 
			
			\hline
			Mnist    & 80  & 97.85$\pm$1.24 & 96.70$\pm$1.71 & 98.90$\pm$0.97
			& \textbf{99.07$\pm$0.94} \\
			0vs5     & 160 & 97.54$\pm$1.46 & 96.84$\pm$1.85 & 98.87$\pm$1.06
			& \textbf{99.22$\pm$0.61} \\
			& 320 & 97.23$\pm$3.34 & 96.88$\pm$0.96 & 98.95$\pm$0.83
			& \textbf{99.27$\pm$0.37} \\
			\hline
			Mnist    & 120 & 94.55$\pm$1.48 & 92.78$\pm$2.11 & 96.02$\pm$1.85
			& \textbf{96.53$\pm$1.49} \\
			0vs3vs5  & 240 & 93.49$\pm$1.07 & 92.88$\pm$1.31 & 94.88$\pm$1.37
			& \textbf{95.37$\pm$0.92} \\
			& 480 & 94.32$\pm$0.81 & 93.37$\pm$1.13 & 95.13$\pm$1.22
			& \textbf{95.54$\pm$0.87} \\
			\hline			
			& 80  & 81.58$\pm$3.10 & 70.83$\pm$4.47 & 82.45$\pm$3.38
			& \textbf{82.65$\pm$3.32} \\
			Splice   & 160 & 84.07$\pm$2.51 & 78.87$\pm$3.01 & 84.87$\pm$2.19
			& \textbf{85.25$\pm$2.06} \\
			& 320 & 84.85$\pm$2.38 & 81.56$\pm$1.99 & 85.94$\pm$1.61
			& \textbf{86.40$\pm$1.59} \\
			\hline
			& 100 & 95.22$\pm$1.30 & 92.47$\pm$1.68 & 96.84$\pm$1.40
			& \textbf{97.29$\pm$1.25} \\
			Gisette  & 200 & 94.38$\pm$1.52 & 92.96$\pm$1.75 & 96.27$\pm$1.53
			& \textbf{96.82$\pm$0.91} \\
			& 300 & 96.11$\pm$0.95 & 95.08$\pm$1.19 & 97.14$\pm$0.87
			& \textbf{97.79$\pm$0.46} \\
			\hline
			USPS     & 120 & 95.42$\pm$1.82 & 94.82$\pm$2.02 & 96.26$\pm$1.33
			& \textbf{97.23$\pm$1.64} \\
			0vs5     & 160 & 96.04$\pm$1.33 & 94.95$\pm$1.70 & 97.03$\pm$1.47
			& \textbf{98.31$\pm$0.82} \\
			& 240 & 96.35$\pm$1.06 & 95.17$\pm$1.16 & 97.24$\pm$0.96
			& \textbf{98.87$\pm$0.74} \\
			\hline
			USPS     & 180 & 93.36$\pm$1.77 & 91.97$\pm$2.00 & 94.86$\pm$1.17
			& \textbf{95.28$\pm$0.96} \\
			0vs3vs5  & 240 & 93.13$\pm$1.38 & 92.01$\pm$1.45 & 94.33$\pm$1.54
			& \textbf{94.96$\pm$1.37} \\
			& 300 & 92.99$\pm$1.35 & 91.81$\pm$1.67 & 93.47$\pm$1.83
			& \textbf{94.05$\pm$1.46} \\
			\hline
			& 60  & 96.50$\pm$1.59 & 97.43$\pm$1.36 & 98.31$\pm$1.10
			& \textbf{98.97$\pm$0.95} \\
			Satimage & 90  & 96.78$\pm$2.72 & 97.31$\pm$1.10 & 98.19$\pm$1.16
			& \textbf{98.71$\pm$1.13} \\
			& 120 & 96.22$\pm$1.91 & 97.23$\pm$1.22 & 98.02$\pm$1.22
			& \textbf{98.53$\pm$1.20} \\
			\hline
		\end{tabular}
	}
	\label{table:ablation_study_one_shot}
\end{table}

\subsection{Configurations and Competing Methods}
The experimental configurations of our EML model in one-shot scenario and some competing methods are detailedly introduced in this subsection.

\begin{figure*}[t]
	\begin{minipage}[t]{0.245\linewidth}
		\centering
		\includegraphics[width=122pt, height=110pt]{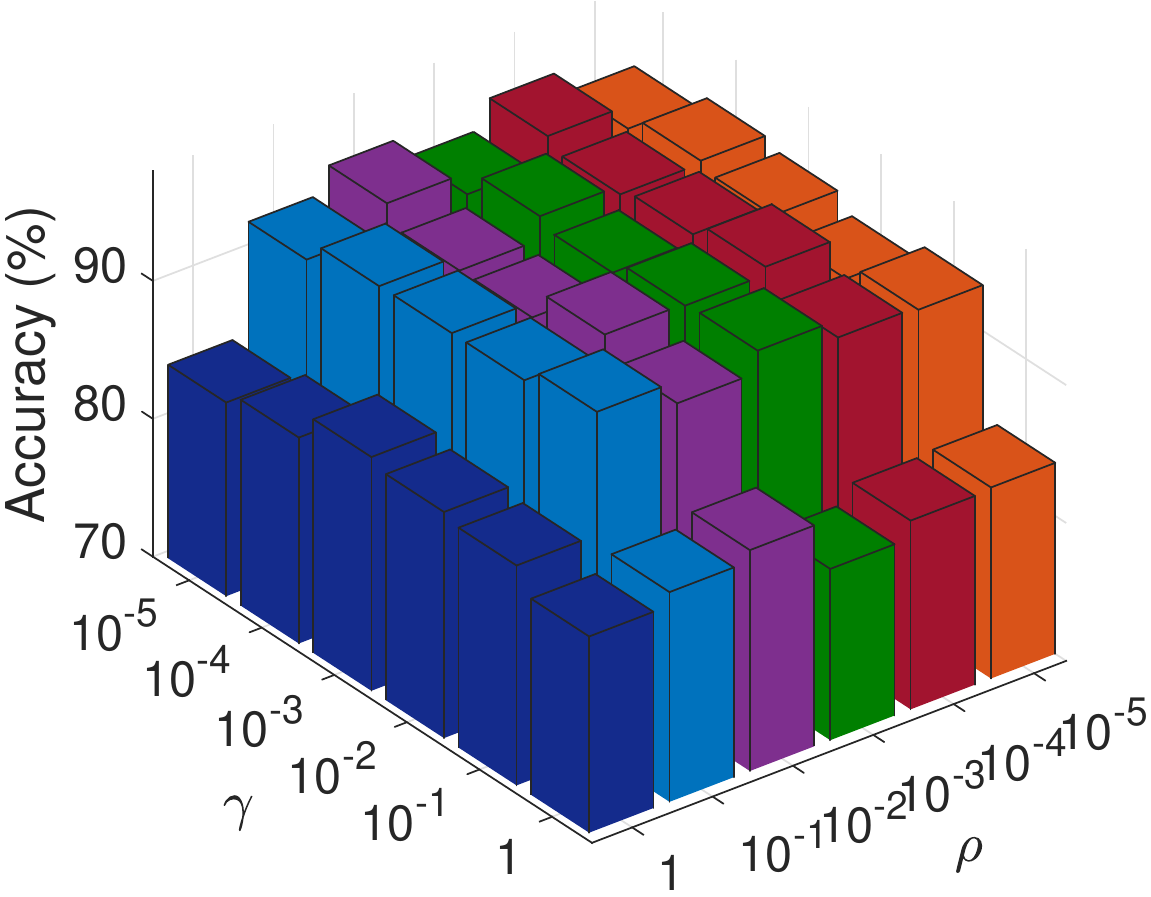}
		(a) Mnist0vs3vs5 ($n_i=480$)
	\end{minipage}
	\begin{minipage}[t]{0.245\linewidth}
		\centering
		\includegraphics[width=122pt, height=110pt]{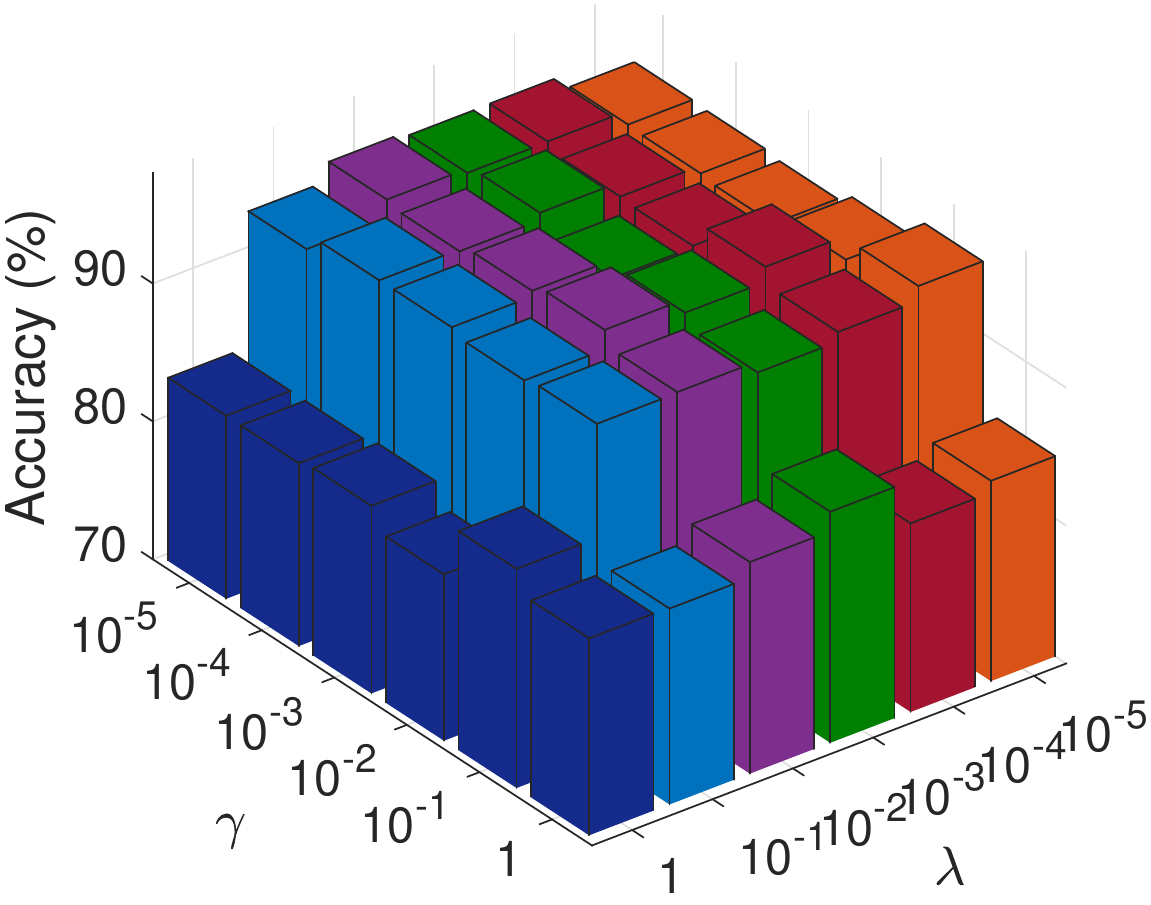}
		(b) Mnist0vs3vs5 ($n_i=480$)
	\end{minipage}
	\begin{minipage}[t]{0.245\linewidth}
		\centering
		\includegraphics[width=122pt, height=110pt]{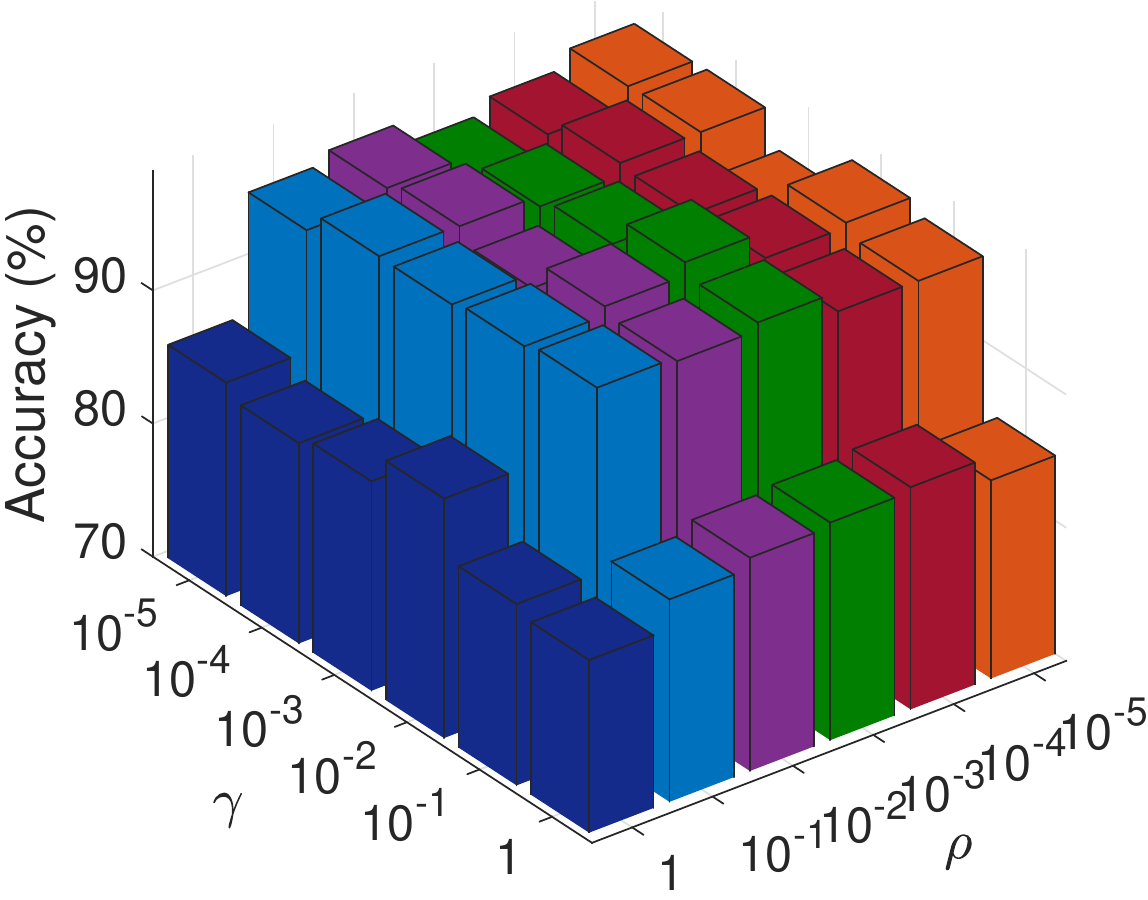}
		(c) USPS0vs5 ($n_i=240$)
	\end{minipage}
	\begin{minipage}[t]{0.245\linewidth}
		\centering
		\includegraphics[width=122pt, height=110pt]{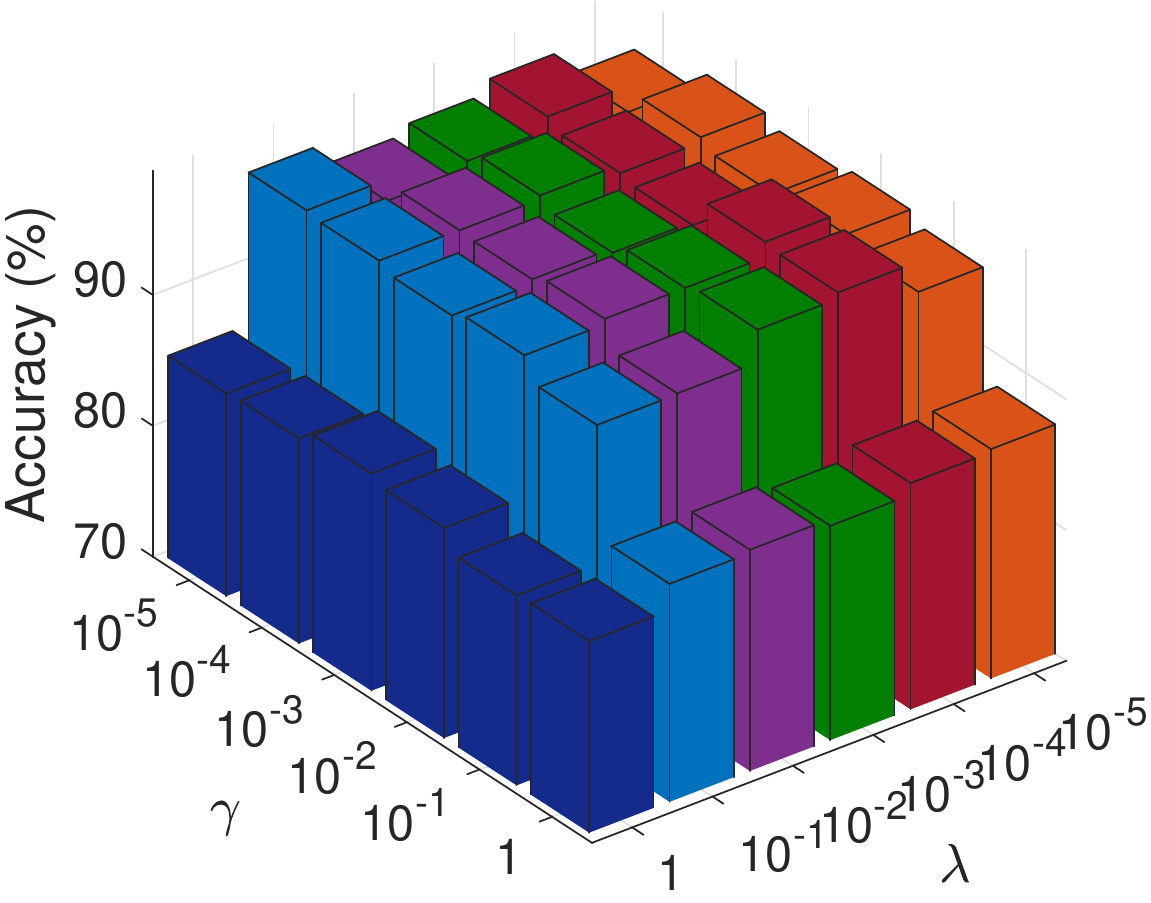}
		(d) USPS0vs5 ($n_i=240$)
	\end{minipage}
	\begin{minipage}[t]{0.245\linewidth}
		\centering
		\includegraphics[width=122pt, height=110pt]{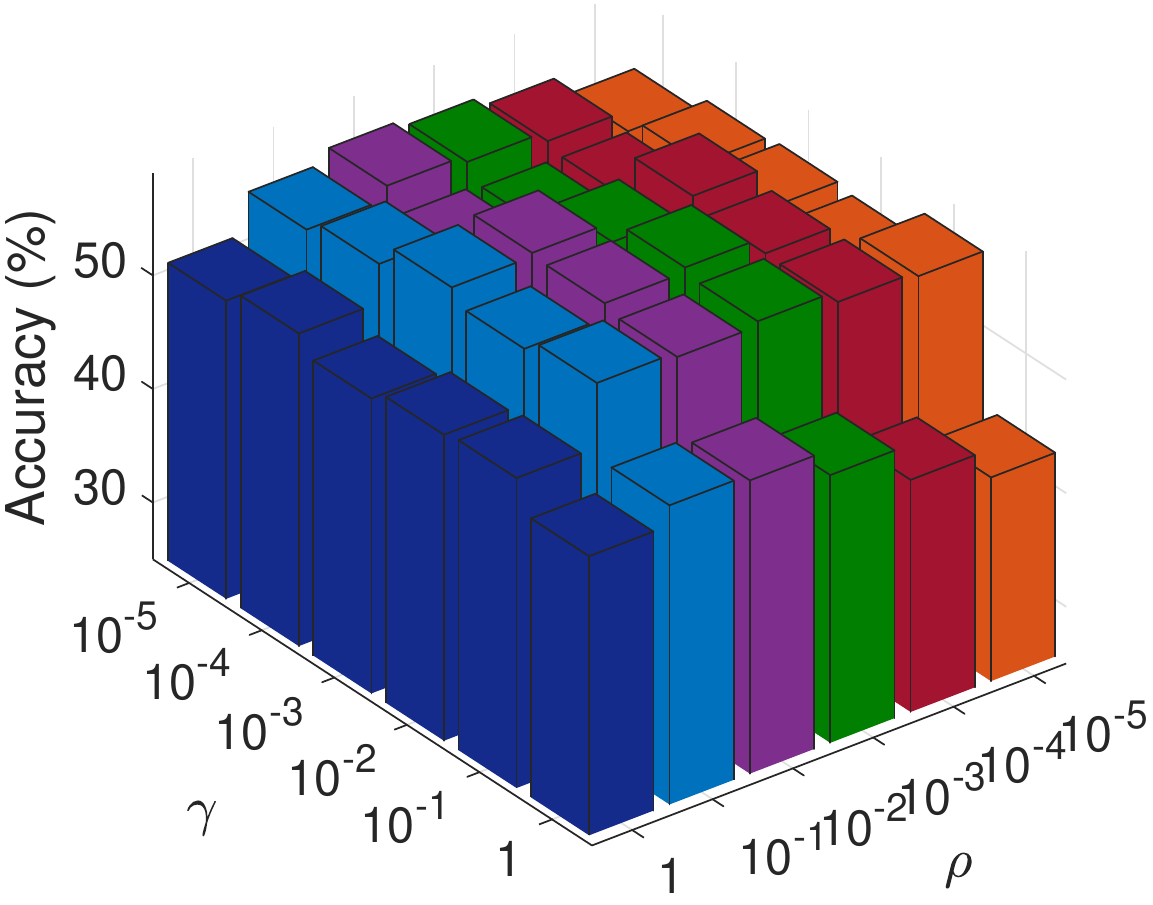}
		(e) EV-Action ($n_i=600$)
	\end{minipage}
	\begin{minipage}[t]{0.245\linewidth}
		\centering
		\includegraphics[width=122pt, height=110pt]{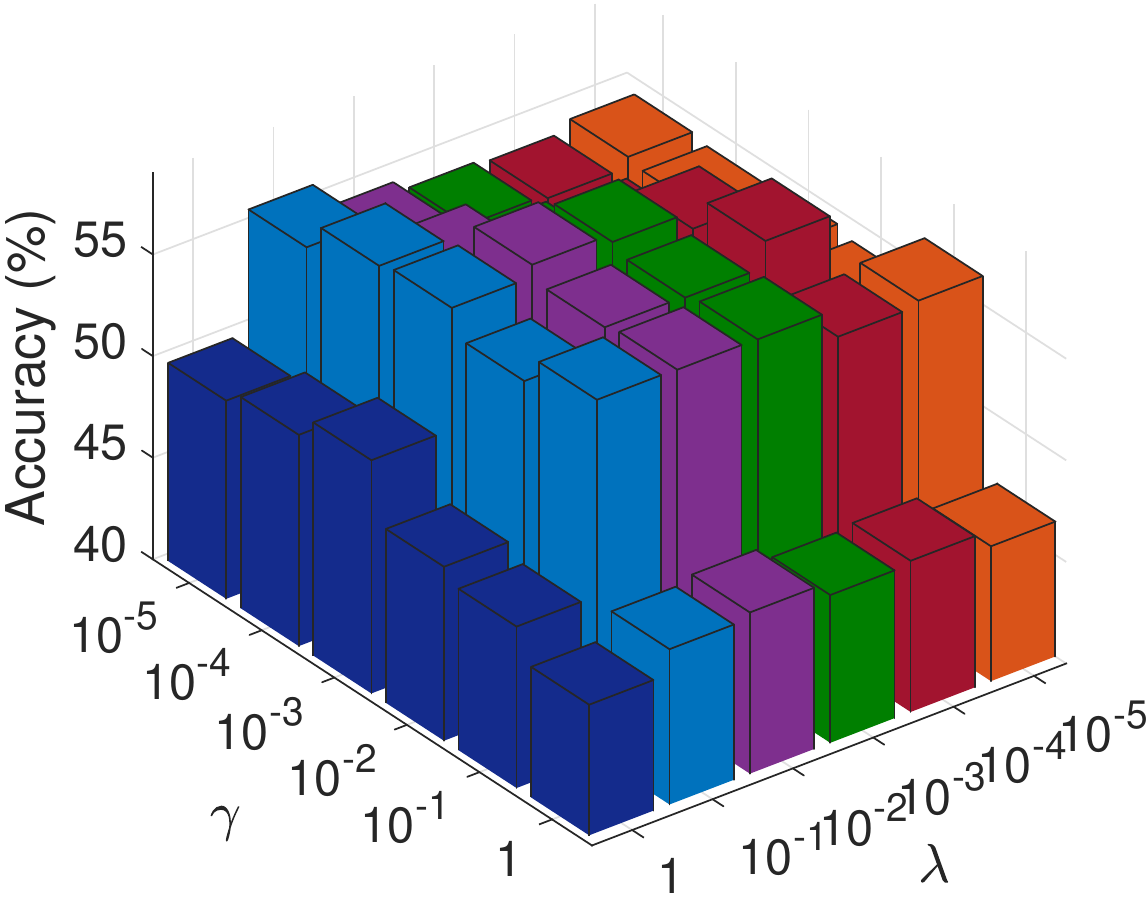}
		(f) EV-Action ($n_i=600$)
	\end{minipage}
	\begin{minipage}[t]{0.245\linewidth}
		\centering
		\includegraphics[width=122pt, height=110pt]{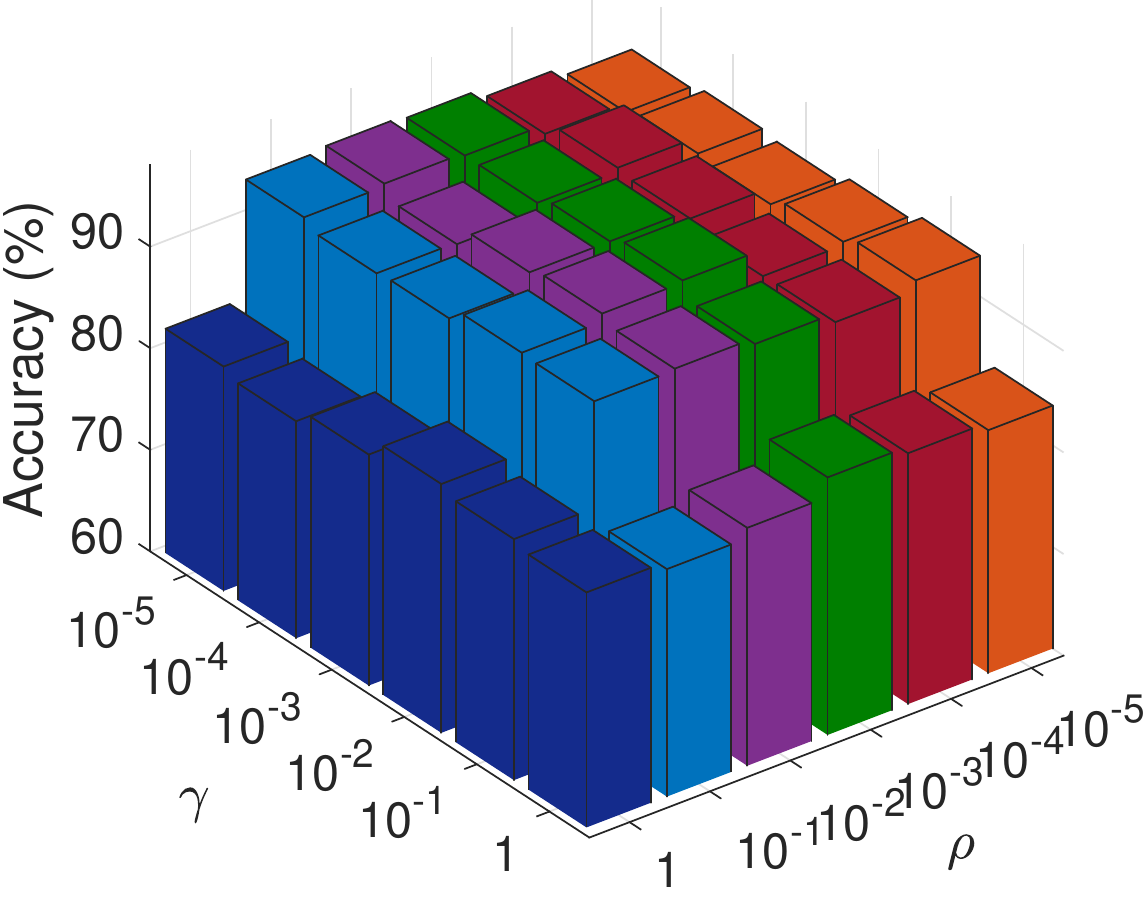}
		(g) PAMAP2 ($n_i=700$)
	\end{minipage}
	\begin{minipage}[t]{0.245\linewidth}
		\centering
		\includegraphics[width=122pt, height=110pt]{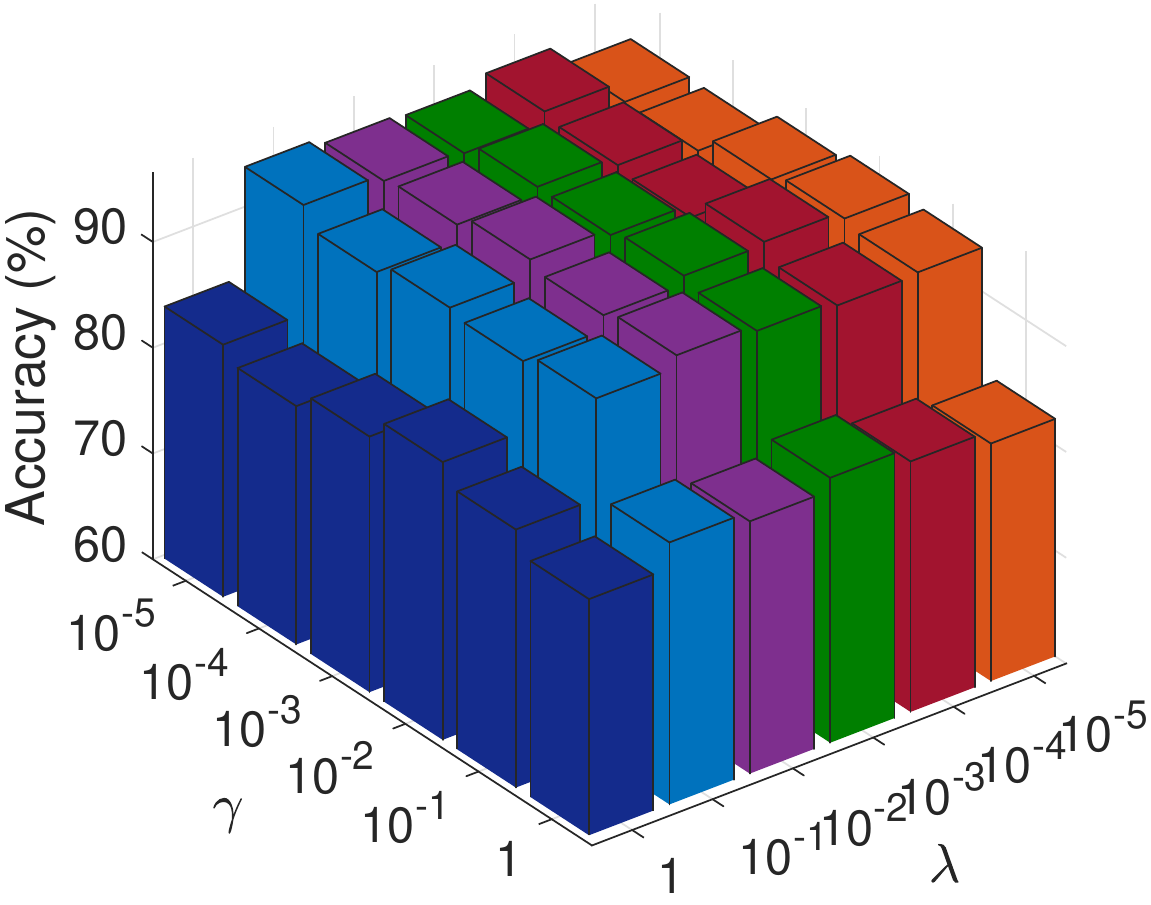}
		(h) PAMAP2 ($n_i=700$)
	\end{minipage}
	\begin{minipage}[t]{0.245\linewidth}
		\centering
		\includegraphics[width=122pt, height=110pt]{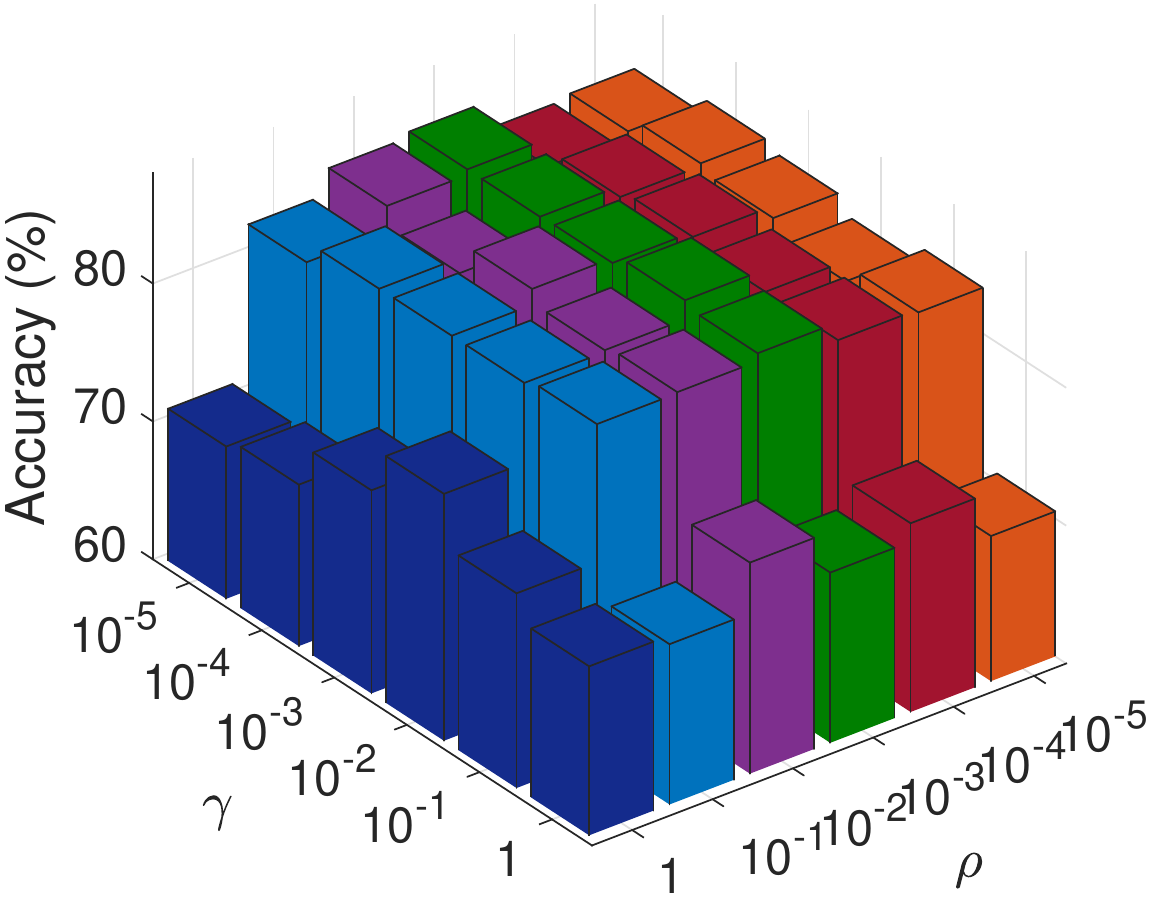}
		(i) Splice ($n_i=600$)
	\end{minipage}
	\begin{minipage}[t]{0.245\linewidth}
		\centering
		\includegraphics[width=122pt, height=110pt]{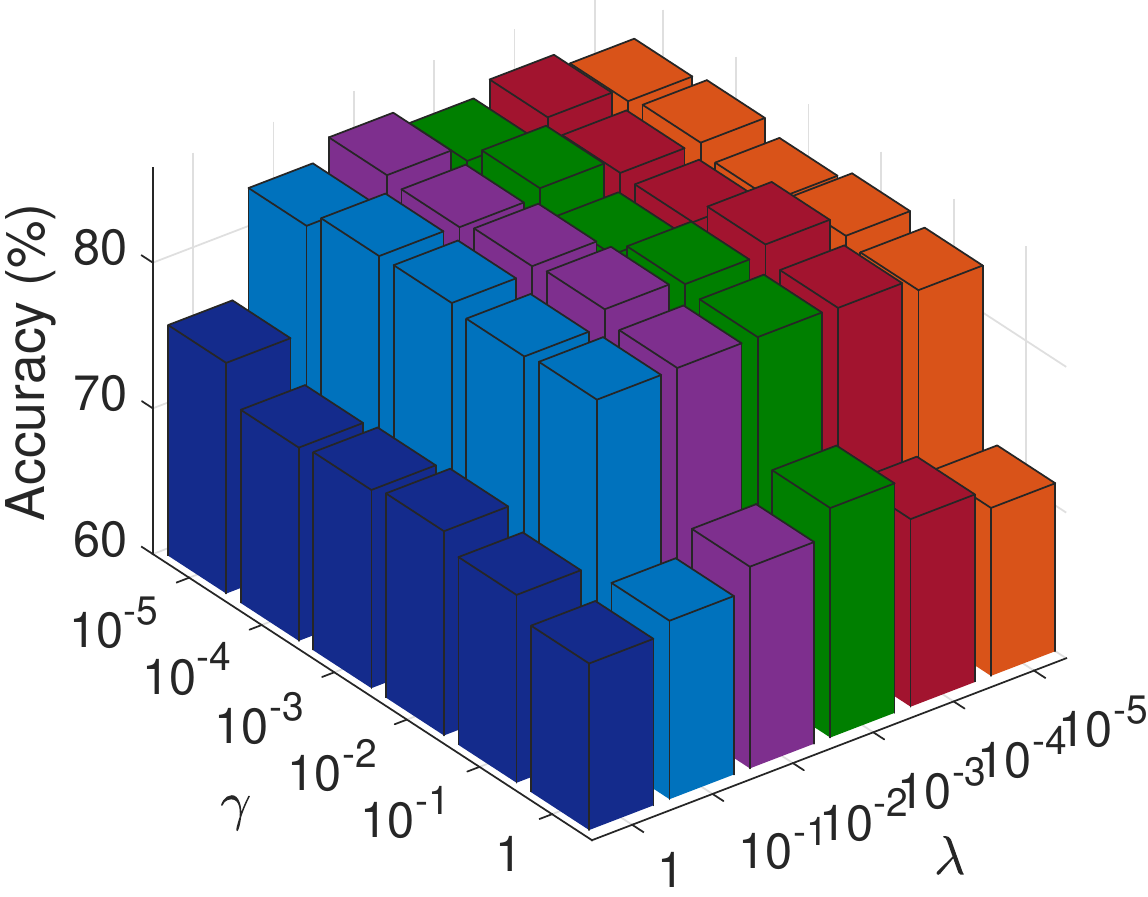}
		(j) Splice ($n_i=600$)
	\end{minipage}
	\begin{minipage}[t]{0.245\linewidth}
		\centering
		\includegraphics[width=122pt, height=110pt]{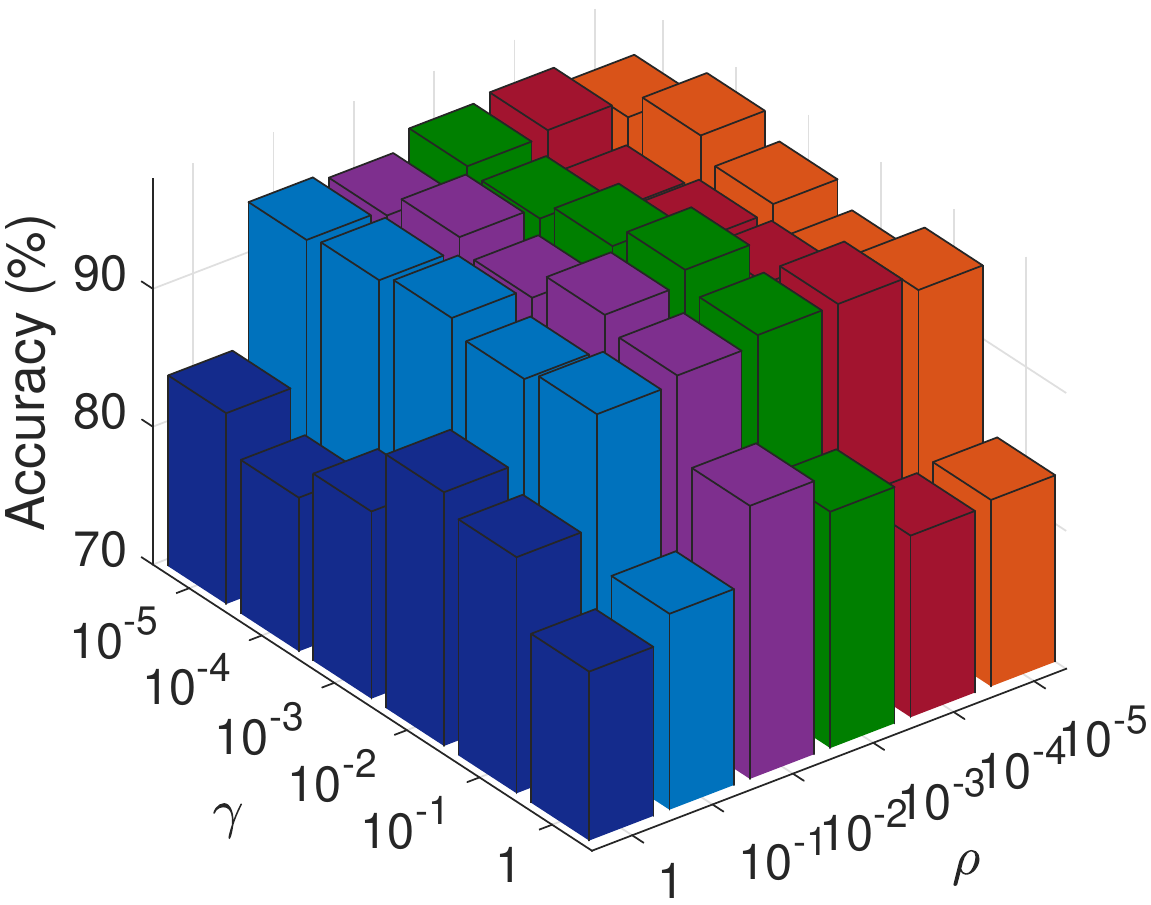}
		(k) Gisette ($n_i=300$)
	\end{minipage}
	\begin{minipage}[t]{0.245\linewidth}
		\centering
		\includegraphics[width=122pt, height=110pt]{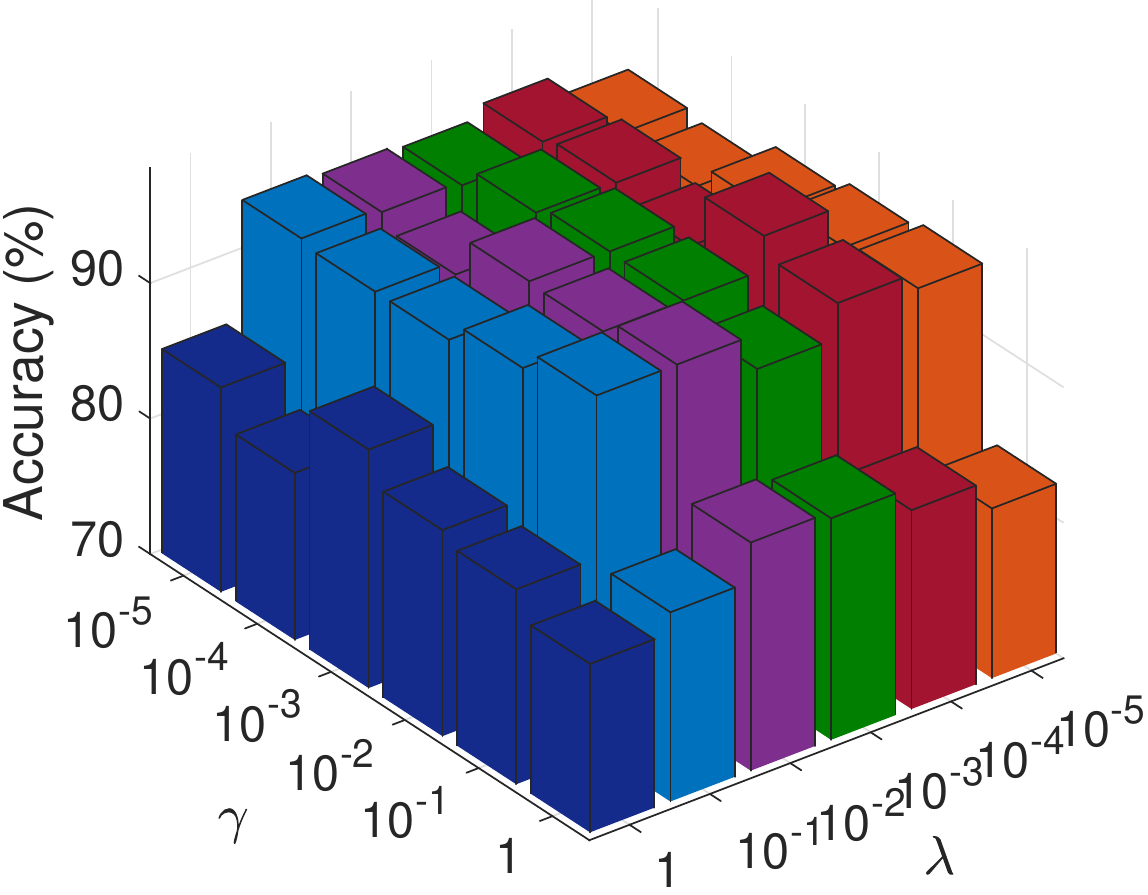}
		(l) Gisette ($n_i=300$)
	\end{minipage}
	\caption{Effect Investigations of hyper-parameters $\{\gamma, \rho\}$ when $\lambda=10^{-4}$ and $\{\gamma, \lambda\}$ when $\rho=10^{-3}$ on Mnist0vs3vs5 (a)(b), USPS0vs5 (c)(d), EV-Action (e)(f), PAMAP2 (g)(h), Splice (i)(j) and Gisette (k)(l) datasets in one-shot scenario.}
	\label{fig:effect_params_one_shot}
\end{figure*}

\begin{figure}[t]
	\centering
	\includegraphics[width =253pt,height =115pt]{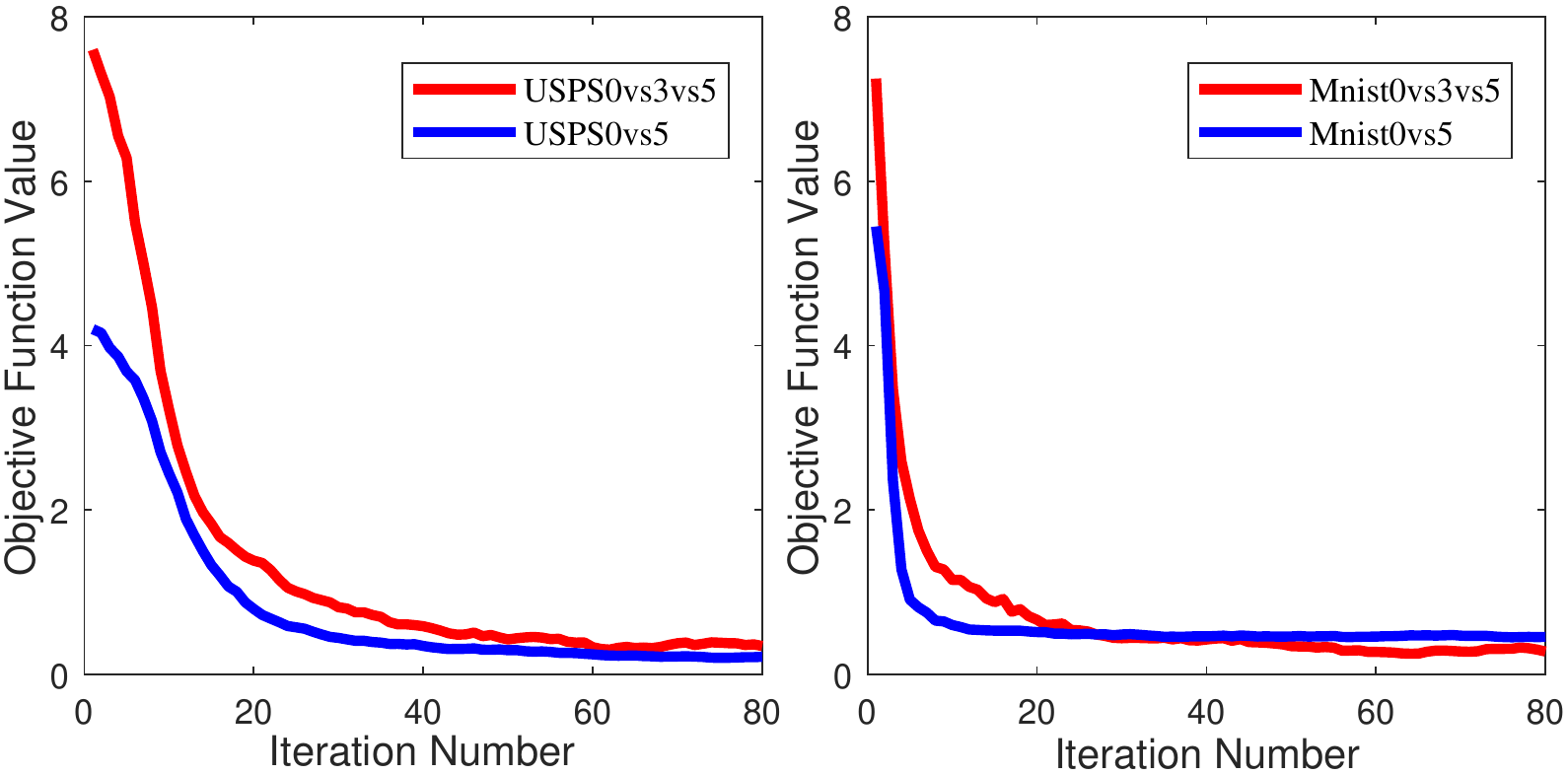}
	\vspace{-20pt}
	\caption{The convergence analysis of our proposed EML model on the USPS (left) and Mnist (right) dataset in one-shot scenario.}  
	\label{fig:converge_analysis}
\end{figure}

\subsubsection{Experimental Configurations} 
As shown in Table~\ref{table:settings_one_shot}, we conduct extensive comparisons on two real-world human motion recognition datasets (\emph{i.e.}, EV-Action \cite{DBLP:journals/corr/abs-1904-12602} and PAMAP2 \cite{6246152}), a large-scale visual recognition dataset (\emph{i.e.}, ImageNet \cite{ILSVRC15}) and five synthetic benchmark datasets\footnote{http://archive.ics.uci.edu/ml/} containing three digit datasets (\emph{i.e.}, Mnist, Gisette and USPS), one DNA dataset (\emph{i.e.}, Splice) and one image dataset (\emph{i.e.}, Satimage). Specifically, EV-Action dataset \cite{DBLP:journals/corr/abs-1904-12602} is a human action dataset with 5300 samples, which consists of 20 common action categories, where 10 actions are finished by single subject and the others are accomplished by the same subjects interacting with other objects. It is a typical application for feature evolution in the real-world, where the features from depth information, RGB image, and human skeleton are respectively regarded as vanished, survived and augmented features. Some example samples about human actions are visualized as Fig.~\ref{fig:examples_human_motion}. 
PAMAP2 \cite{6246152} is composed of 18 activities performed by 9 different subjects wearing three inertial measurement units (IMU) and a heart rate monitor. We only utilize the data information from IMU in our experiments, due to the large missing values collected from the heart rate monitor. Each IMU contains one gyroscope, two accelerometers, one magnetometer, where the features from them are regarded as vanished, survived and augmented features, respectively. Moreover, ImageNet \cite{ILSVRC15} including 1000 different categories is a large-scale challenging visual recognition dataset, where each of 1000 classes has roughly 1000 samples. We utilize ResNet \cite{He2015} as feature extractor to obtain 2048-dimension feature representations for ImageNet \cite{ILSVRC15}.

\begin{table*}[t]
	\centering
	\setlength{\tabcolsep}{0.8mm}
	\caption{Computational time in terms of minutes: mean and standard errors averaged over fifty random runs in one-shot scenario.}
	\scalebox{0.99}{
		\begin{tabular}{|c|cccccccccc|}
			\hline
			Dataset & Pegasos \cite{Shalev-Shwartz2011} & OPMV \cite{Zhu2015OnePassML} & TCA \cite{5640675} & BDML \cite{Xu:2018:BDM:3327144.3327333} & OPML \cite{LI2018302} & CDML \cite{Chen2019CurvilinearDM} & OPIDe \cite{Hou2018OnePassLW} & OPID \cite{Hou2018OnePassLW} & FIRF \cite{8410016} & Ours  \\
			\hline
			
			EV-Action ($n_i = 500$) & 27.11$\pm$0.04 & 28.58$\pm$0.04 & 37.72$\pm$0.09 & 36.18$\pm$0.18 & \textbf{22.95$\pm$0.07} & 37.42$\pm$0.33 & 26.47$\pm$0.09 & 26.33$\pm$0.14 & 23.04$\pm$0.11 & 25.48$\pm$0.06 \\
			
			Mnist0vs5 ($n_i = 80$) & 6.18$\pm$0.07 & 7.45$\pm$0.06 & 14.96$\pm$0.12 & 16.27$\pm$0.10 & \textbf{3.84$\pm$0.04} & 16.58$\pm$0.19 & 5.13$\pm$0.11 & 4.95$\pm$0.07 & 3.89$\pm$0.07 & 4.68$\pm$0.05 \\
			
			USPS0vs5 ($n_i = 120$) & 3.16$\pm$0.02 & 4.75$\pm$0.10 & 11.93$\pm$0.04 & 13.06$\pm$0.12 & \textbf{1.24$\pm$0.03} & 13.42$\pm$0.21 & 1.93$\pm$0.05 & 1.87$\pm$0.06 & 1.32$\pm$0.05 & 1.53$\pm$0.08 \\
			
			Gisette ($n_i = 100$) & 40.52$\pm$0.03 & 41.06$\pm$0.16 &51.28$\pm$0.04 & 49.73$\pm$0.14 & \textbf{35.26$\pm$0.03} & 49.80$\pm$0.17 & 38.24$\pm$0.08 & 39.15$\pm$0.05 & 35.54$\pm$0.15 & 37.57$\pm$0.11 \\
			
			Satimage ($n_i = 120$) & 2.64$\pm$0.05 & 3.08$\pm$0.10 & 10.23$\pm$0.07 & 11.47$\pm$0.09 & \textbf{0.52$\pm$0.02} & 11.62$\pm$0.14 & 0.66$\pm$0.04 & 0.71$\pm$0.08 & 0.55$\pm$0.04 & 0.68$\pm$0.03 \\
			
			PAMAP2 ($n_i = 360$) & 9.84$\pm$0.04 & 9.27$\pm$0.06 & 16.74$\pm$0.15 & 18.05$\pm$0.18 & \textbf{4.69$\pm$0.13} & 18.32$\pm$0.09 & 7.84$\pm$0.09 & 7.63$\pm$0.07 & 4.81$\pm$0.07 & 6.45$\pm$0.06 \\
			
			\hline	
			
		\end{tabular}
	}		
	\label{tab:Computational_time}
\end{table*}

For a fair comparison, as presented in Table~\ref{table:settings_one_shot}, we adopt the same experimental settings with \cite{Hou2018OnePassLW} in one-shot and multi-shot cases, which are elaborated as follows: 
\begin{itemize}
	\setlength{\itemsep}{1pt}
	\setlength{\parsep}{0pt}
	\setlength{\parskip}{0pt}
	\item The number of streaming data in each batch is same, \emph{i.e.}, $n_i = n_{r+1} = {n_{r+2}}  (i\in\{1,2,...,r\})$, and the sample number in each class is equal for all training and testing batches. 
	
	\item In T-stage, the total number of training data is fixed and the sample number in each batch is varied. In the light of this, the number of training and evaluation samples also varies in the last evaluation phase. 
	
	\item We allocate the first $d_v$ features, the next $d_s$ features and and the rest of features as vanished features,  survived features and new augmented features, respectively. The first and last quarters are corresponding vanished and augmented features in our experiments.
	
	\item The experimental performance in each run may have slightly difference due to the influence of computer system and simulation environment, even though we run each experiment under the same experimental settings. To circumvent the randomness effect of experimental performance, all experimental results are the averaged results over fifty random runs, which is more convincing to illustrate the superiority of our EML model.
\end{itemize}

\subsubsection{Competing Methods} 
We validate the superior performance of our EML model by comparing it with the following competing methods: One-pass \textbf{Pegasos} \cite{Shalev-Shwartz2011} assumes that the vanished and augmented features are available in different feature evolution stages; \textbf{OPMV} \cite{Zhu2015OnePassML} regards the features in T-stage and I-stage as the first and second views; \textbf{TCA} \cite{5640675} assumes that the streaming samples in T-stage and I-stage are drawn from the source and target distributions; \textbf{BDML} \cite{Xu:2018:BDM:3327144.3327333},  \textbf{OPML} \cite{LI2018302} and \textbf{CDML} \cite{Chen2019CurvilinearDM} are the representative metric learning methods, which only utilize the samples with the augmented features, and ignore the previous vanished features; As for the feature evolution approaches, \textbf{OPID} and \textbf{OPIDe} \cite{Hou2018OnePassLW} propose an one-pass incremental and decremental model for feature evolution. 
\textbf{FIRF} \cite{8410016} designs a feature incremental random forest framework to tackle the emergence of new sensors (\emph{i.e.}, new augmented features) in a dynamic environment.

\subsection{Experiments in One-shot Scenario}
In this subsection, we introduce the comprehensive experimental analysis, ablation studies, effects of hyper-parameters and convergence investigation of our proposed EML model in one-shot scenario, followed by computational costs of optimization complexity.

\subsubsection{Experimental Analysis}
The experimental results for one-shot scenario are presented in Table~\ref{table:results_one_shot}. From the presented performance, we have the following observations: 1) Although our proposed model has no access to the vanished features in T-stage, both transforming and inheriting strategies could efficiently exploit useful information of vanished feature and expand it into new augmented features in I-stage. 2) Our proposed EML model could be successfully applied to both high-dimensional (\emph{e.g.}, EV-Action, Gisette and ImageNet) and  low-dimensional (\emph{e.g.}, Satimage and Splice) feature evolution, which are the challenging tasks to explore the intrinsic data structure and informative knowledge using the existing features; 3) When we utilize the learned distance matrix in T-stage to assist the training procedure in I-stage, the evaluation performance of our proposed model increases significantly, even though the training samples in I-stage are relatively rare, \emph{i.e.}, $n_i$ contains a small number of samples in I-stage. 4) Our EML model performs better than OPID and OPIDe \cite{Hou2018OnePassLW}, since T-stage could explore important information from vanished features, and I-stage efficiently inherits the metric performance from T-stage to take advantage of new augmented features.

\begin{table}[t]
	\centering
	\setlength{\tabcolsep}{5.4mm}
	\caption{Effect investigation of low rank constraint of our EML model in one-shot scenario.}
	\scalebox{1.0}{
		\begin{tabular}{|c|c|cc|}
			\hline
			Dataset & $n_i$ & Ours-woLR & Ours \\
			\hline	
			
			& 500  & 57.61$\pm$0.56 & \textbf{58.33$\pm$0.76} \\ 
			EV-Action  & 600 & 57.46$\pm$0.89 & \textbf{57.94$\pm$0.88} \\ 
			& 700 & 57.35$\pm$1.34 & \textbf{58.12$\pm$0.86} \\ 
			
			\hline
			Mnist    & 120 & 96.18$\pm$1.36 & \textbf{96.53$\pm$1.49} \\
			0vs3vs5  & 240 & 94.62$\pm$1.04	& \textbf{95.37$\pm$0.92} \\
			& 480 &  95.20$\pm$1.01 & \textbf{95.54$\pm$0.87} \\
			
			\hline			
			& 80  & 82.16$\pm$0.94 & \textbf{82.65$\pm$3.32} \\
			Splice   & 160  & 84.69$\pm$1.93 & \textbf{85.25$\pm$2.06} \\
			& 320 & 85.76$\pm$1.45 & \textbf{86.40$\pm$1.59} \\

			\hline
			& 100  & 96.65$\pm$1.23 & \textbf{97.29$\pm$1.25} \\
			Gisette  & 200  & 96.40$\pm$1.16 & \textbf{96.82$\pm$0.91} \\
			& 300  & 97.08$\pm$0.75 & \textbf{97.79$\pm$0.46} \\ 
						
			\hline
			USPS     & 180 & 94.59$\pm$1.05 & \textbf{95.28$\pm$0.96} \\
			0vs3vs5  & 240  & 94.51$\pm$1.27 & \textbf{94.96$\pm$1.37} \\
			& 300 & 93.29$\pm$1.18 & \textbf{94.05$\pm$1.46} \\
			
			\hline
			& 60  & 98.42$\pm$1.22 & \textbf{98.97$\pm$0.95} \\
			Satimage & 90  & 98.32$\pm$1.30 & \textbf{98.71$\pm$1.13} \\
			& 120 & 98.11$\pm$1.05 & \textbf{98.53$\pm$1.20} \\
			\hline
		\end{tabular}
	}
	\label{table:effect_low_rank_one_shot}
\end{table}

\subsubsection{Ablation Studies} 
To verify the effectiveness of our EML model, we intend to research the effects of different components of our model, \emph{i.e.}, training without T-stage (denoted as Ours-woT), training without I-stage (denoted as Ours-woI) and training without the Wasserstein distance metric (denoted as Ours-woW). The performance of Ours-woW is evaluated under the metric of Mahalanobis distance. From the presented results in Table~\ref{table:ablation_study_one_shot}, our proposed EML model has the best performance when both transforming and inheriting strategies work together to tackle incremental and decremental features via the Wasserstein distance metric, which validates the reasonable design of our proposed model. Compared with other metric distances (\emph{e.g.}, Mahalanobis distance), the smoothed Wasserstein distance could better mine the similarity relationships between the heterogeneous and complex streaming samples, since the evolving features are not strictly aligned in different stages. Both T-stage and I-stage play an essential role in tackling instance and feature evolutions simultaneously.

\subsubsection{Effects of Hyper-Parameters} 
In this subsection, as shown in Fig.~\ref{fig:effect_params_one_shot}, we introduce extensive parameter experiments on several representative datasets (Mnist0vs3vs5, USPS0vs5, EV-Action, PAMAP2, Splice and Gisette) as the examples to investigate the effects of hyper-parameters $\{\gamma, \lambda, \rho\}$ in one-shot scenario. Specifically, the experimental performances of our proposed model are averaged over fifty random repetitions by empirically tuning $\{\gamma, \lambda, \rho\}$ in a wide selection range of $\{10^{-5}, 10^{-4}, 10^{-3}, 10^{-2}, 10^{-1}, 1\}$ to choose the optimal values of hyper-parameters. When fixing $\lambda$ as $10^{-4}$, we investigate the effects of $\{\gamma, \rho\}$, and introduce the hyper-parameter influence of $\{\gamma, \lambda\}$ when $\rho=10^{-3}$. From the performance depicted in Fig.~\ref{fig:effect_params_one_shot}, we could observe our EML model has stable prediction performance over the wide selection range of different hyper-parameters. Moreover, when $\gamma=10^{-2}, \rho=10^{-3}$ and $\lambda=10^{-4}$, our EML model performs the best prediction performance on most benchmark dataset, except for Mnist0vs3vs5 dataset performing the best when $\gamma=10^{-2}, \rho=10^{-4}$ and $\lambda=10^{-4}$.

\begin{table*}[t]
	\centering
	\setlength{\tabcolsep}{1.35mm}
	\caption{Comparisons between our model and state-of-the-art methods in terms of accuracy (\%) on seven datasets: mean and standard errors averaged over fifty random runs in multi-shot scenario for Task I. Models with the best performance are bolded.}
	\scalebox{0.95}{
		\begin{tabular}{|c|c|cccccc|ccc|c|}
			\hline
			Dataset & $n_i$ & Pegasos \cite{Shalev-Shwartz2011} & OPMV \cite{Zhu2015OnePassML} & TCA \cite{5640675} & BDML \cite{Xu:2018:BDM:3327144.3327333} & OPML \cite{LI2018302} & CDML \cite{Chen2019CurvilinearDM} & OPIDe \cite{Hou2018OnePassLW} & OPID \cite{Hou2018OnePassLW} & FIRF \cite{8410016} & Ours \\
			\hline
			& 500  & 54.25$\pm$1.42 & 53.60$\pm$1.53 & 50.63$\pm$1.89
			& 53.26$\pm$1.18
			& 51.36$\pm$1.48 & 52.77$\pm$0.69 & 54.61$\pm$1.53 & 54.27$\pm$1.24 & 54.16$\pm$0.57 & \textbf{55.93$\pm$1.04} \\
			EV-Action & 600 & 55.59$\pm$1.27 & 53.72$\pm$1.66 & 51.83$\pm$1.96
			& 53.74$\pm$0.82
			& 53.11$\pm$1.36 & 52.74$\pm$1.28 & 54.43$\pm$1.15 & 53.46$\pm$1.92 & 53.84$\pm$1.07 & \textbf{56.74$\pm$1.35} \\
			& 700 & 54.19$\pm$1.28 & 53.52$\pm$1.74  & 51.95$\pm$1.36
			& 53.84$\pm$0.59
			& 52.54$\pm$1.83 & 53.61$\pm$0.94 & 54.38$\pm$1.19 & 54.62$\pm$1.02 & 55.04$\pm$1.30 & \textbf{56.19$\pm$0.77} \\	
			
			\hline
			Mnist     & 80  & 97.50$\pm$1.82 & 88.75$\pm$4.99 & 93.14$\pm$1.87
			& 92.25$\pm$2.20 & 93.92$\pm$3.22 & 92.74$\pm$2.03 & 95.70$\pm$2.17
			& 95.92$\pm$2.23 & 94.37$\pm$2.16 & \textbf{98.54$\pm$1.08} \\
			0vs5      & 160 & 97.56$\pm$1.28 & 90.75$\pm$3.02 & 91.78$\pm$2.43
			& 95.70$\pm$1.26 & 94.34$\pm$1.54 & 95.87$\pm$1.85 & 95.53$\pm$1.61
			& 95.29$\pm$1.80 & 94.16$\pm$1.85 & \textbf{98.61$\pm$0.57} \\
			~ 		 & 320 & 97.61$\pm$0.90 & 92.72$\pm$1.76 & 90.35$\pm$2.67
			& 96.06$\pm$0.98 & 95.20$\pm$0.96 & 95.74$\pm$1.73 & 95.22$\pm$1.33
			& 95.04$\pm$1.39 & 94.61$\pm$1.17 & \textbf{98.73$\pm$0.64} \\
			\hline 			
			~  	 	 & 100 & 91.58$\pm$2.87 & 86.24$\pm$4.76 & 91.28$\pm$2.56
			& 90.48$\pm$3.29 & 89.87$\pm$3.62 & 90.26$\pm$2.71 & 95.08$\pm$2.35
			& 94.36$\pm$1.88 & 92.88$\pm$2.06 & \textbf{96.12$\pm$1.18} \\
			Gisette   & 200 & 90.68$\pm$1.79 & 88.41$\pm$3.00 & 90.92$\pm$2.84
			& 90.69$\pm$2.73 & 92.22$\pm$2.03 & 91.23$\pm$1.89 & 94.88$\pm$1.39
			& 93.81$\pm$1.50 & 93.14$\pm$1.83 & \textbf{95.94$\pm$1.72} \\
			~		 & 300 & 91.18$\pm$1.13 & 89.42$\pm$2.27 & 92.13$\pm$2.14
			& 92.52$\pm$1.71 & 91.83$\pm$1.61 & 92.06$\pm$1.36 & 94.65$\pm$1.12
			& 93.91$\pm$1.45 & 93.08$\pm$1.26 & \textbf{95.71$\pm$1.68} \\
			\hline
			USPS      & 120 & 97.48$\pm$0.12 & 94.95$\pm$2.46 & 92.87$\pm$2.16
			& 96.08$\pm$1.36 & 93.83$\pm$2.13 & 95.84$\pm$1.71 & 94.77$\pm$1.62
			& 94.61$\pm$1.69 & 93.92$\pm$1.53 & \textbf{98.57$\pm$0.94} \\
			0vs5      & 160 & 97.56$\pm$0.19 & 95.17$\pm$2.18 & 93.05$\pm$1.94
			& 96.24$\pm$1.55 & 94.21$\pm$1.66 & 95.49$\pm$1.33 & 94.13$\pm$1.54
			& 94.51$\pm$1.30 & 93.75$\pm$1.28 & \textbf{98.68$\pm$0.65} \\
			~         & 240 & 97.37$\pm$0.41 & 95.58$\pm$1.06 & 92.72$\pm$2.33
			& 97.57$\pm$0.67 & 94.62$\pm$1.25 & 95.84$\pm$2.03 & 93.92$\pm$1.47
			& 93.62$\pm$1.16 & 92.64$\pm$1.09 & \textbf{98.39$\pm$0.72} \\
			\hline 		
			USPS      & 180 & 92.03$\pm$1.57 & 89.22$\pm$3.21 & 90.86$\pm$2.87
			& 90.91$\pm$1.81 & 88.61$\pm$2.84 & 89.73$\pm$2.43 & 84.05$\pm$2.27
			& 83.34$\pm$2.36 & 82.94$\pm$2.36 & \textbf{93.11$\pm$1.87} \\
			0vs3vs5   & 240 & 90.90$\pm$1.40 & 89.13$\pm$2.05 & 90.24$\pm$2.93
			& 91.98$\pm$2.04 & 89.62$\pm$2.17 & 90.62$\pm$1.87 & 84.68$\pm$1.73
			& 84.49$\pm$1.92 & 83.75$\pm$1.81 & \textbf{93.23$\pm$1.58} \\
			~	     & 300 & 90.48$\pm$1.24 & 89.52$\pm$1.59 & 89.85$\pm$3.16
			& 92.61$\pm$1.23 & 89.46$\pm$1.80 & 90.84$\pm$1.56 & 83.25$\pm$1.61
			& 83.17$\pm$1.66 & 84.23$\pm$1.42 & \textbf{93.13$\pm$1.55} \\
			\hline
			& 10000 & 52.76$\pm$1.55 & 48.11$\pm$2.30 & 47.82$\pm$3.06 & 49.04$\pm$2.83 & 49.74$\pm$2.49 & 49.43$\pm$2.59 & 50.42$\pm$1.57 & 50.33$\pm$2.26 & 48.42$\pm$2.01 & \textbf{53.92$\pm$1.62} \\
			ImageNet & 12000 & 53.81$\pm$1.44 & 48.15$\pm$2.06 & 48.82$\pm$2.47 & 50.17$\pm$2.42 & 50.63$\pm$1.42 & 49.95$\pm$2.51 & 52.36$\pm$2.13 & 53.09$\pm$2.28 & 51.64$\pm$1.93 & \textbf{54.95$\pm$2.17} \\
			& 14000 & 55.82$\pm$2.05 & 48.93$\pm$3.28 & 48.36$\pm$2.66 & 50.93$\pm$2.87 & 51.62$\pm$2.62 & 50.51$\pm$2.18 & 53.62$\pm$1.83 & 54.96$\pm$2.10 & 51.63$\pm$2.14 & \textbf{57.03$\pm$1.92} \\
			\hline
			
			& 600 & 89.07$\pm$1.25 & 86.71$\pm$1.22 & 82.83$\pm$2.17 & 85.82$\pm$1.55 & 83.06$\pm$1.39 & 84.28$\pm$1.93 & 90.42$\pm$1.25 & 90.37$\pm$1.52 & 91.28$\pm$0.66 & \textbf{92.75$\pm$0.97} \\
			PAMAP2 & 700 & 88.47$\pm$1.58 & 87.05$\pm$1.42 & 
			83.23$\pm$1.69 & 85.52$\pm$1.74 & 82.12$\pm$1.26 & 85.68$\pm$1.64 & 88.63$\pm$1.33 & 90.18$\pm$0.94 & 90.57$\pm$1.24 & \textbf{92.51$\pm$0.92} \\
			& 800 & 88.74$\pm$1.03 & 88.05$\pm$1.64 & 82.19$\pm$2.26 & 85.48$\pm$1.94 & 84.95$\pm$1.68 & 86.26$\pm$1.33 & 90.87$\pm$0.92 & 91.73$\pm$1.15 & 91.62$\pm$0.81 & \textbf{92.38$\pm$1.03} \\
			\hline
		\end{tabular}
	}
	\label{table:results_multi_shot_task_1}
\end{table*}

\begin{figure*}[t]
	\centering
	\includegraphics[width =470pt,height =225pt]{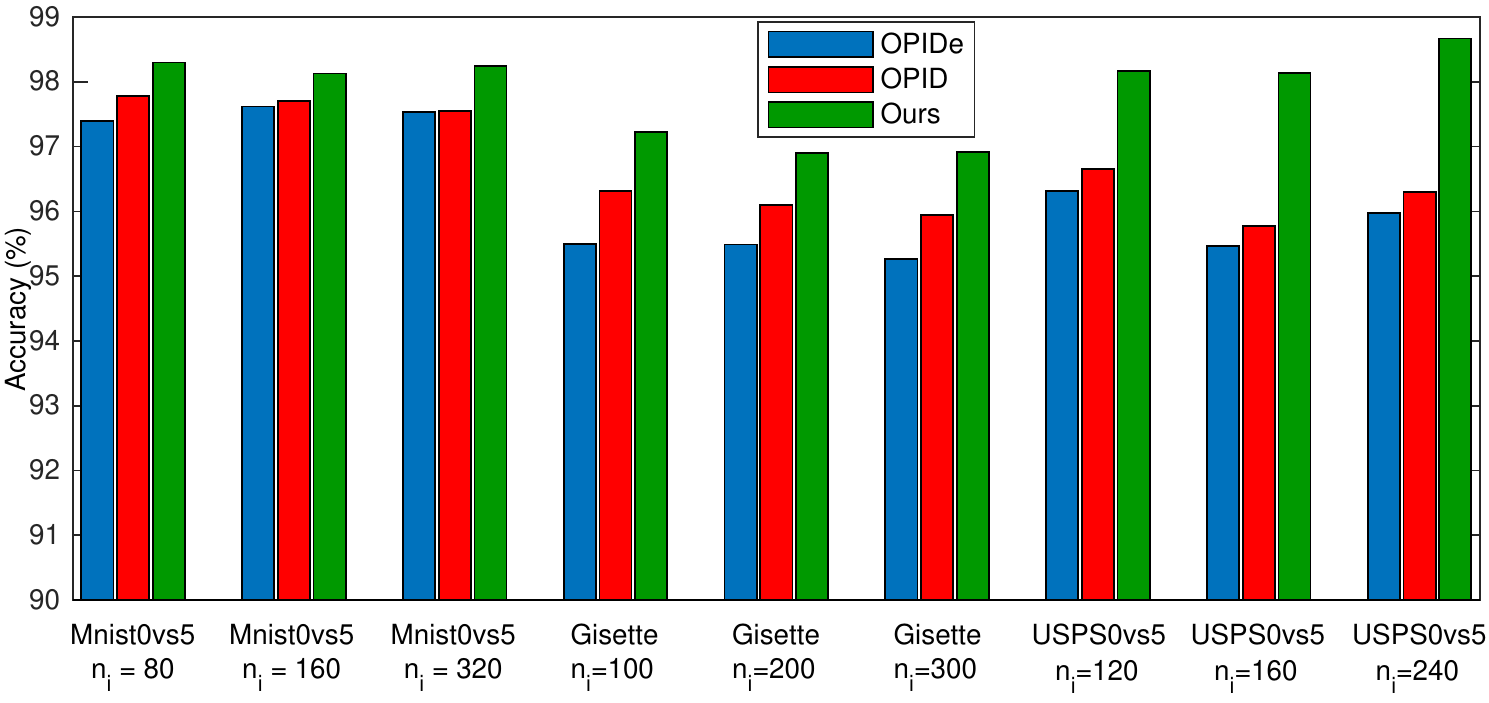}
	\vspace{-10pt}
	\caption{Comparisons in terms of accuracy (\%) on three datasets: mean and standard errors averaged over fifty random runs in multi-shot scenario for Task II.}
	\label{fig:results_multi_shot_task_2}
\end{figure*}

\subsubsection{Convergence Investigations}
The convergence condition of \textbf{Algorithm 1} is depending on the little change (we set it as $2.5\times10^{-5}$) in the consecutive objective function values, and Fig.~\ref{fig:converge_analysis} depicts the convergence curves of our EML model on Mnist and USPS datasets. From the presented results in Fig.~\ref{fig:converge_analysis}, we notice that our EML model could converge asymptotically to a stable value with respect to the objective function value after a few iterations. Furthermore, it validates that our proposed optimization algorithm could efficiently achieve stable performance with appropriate convergence condition.

\subsubsection{Computational Costs} 
Table~\ref{tab:Computational_time} presents the computational costs (\emph{i.e.}, optimization time) of our proposed EML model and other competing methods. From the reported results, we have the following conclusions: 1) Our model is computationally efficient in an online manner for real-world applications since $n_p, n_q, n_k \ll n_i$ and $k$ is often a small value when compared with the feature dimension and the sample number. 
2) The computational time costs (by the minute) of our model are less than other competing methods about 0.67$\sim$13.71 minutes on most experimental datasets except for OPML \cite{LI2018302}, since OPML only takes advantage of training samples in I-stage for optimization procedure.

\subsubsection{Effect Investigation of Low Rank Constraint}
This subsection investigates the effectiveness of low rank regularizer in our proposed EML model, as introduced in Table~\ref{table:effect_low_rank_one_shot}. We substitute the low rank constraint with Frobenius norm, and denote its classification performance as Ours-woLR. The presented results in Table~\ref{table:effect_low_rank_one_shot} clearly demonstrates that the performance of our proposed EML model degrades about $0.42\%\sim0.72\%$ in terms of accuracy, when the low rank constraint is abandoned. It illustrates that our EML model could effectively explore the intrinsic low rank structure of heterogeneous samples for different evolving features by incorporating with the low rank regularizer.

\subsection{Experiments in Multi-shot Scenario}
This subsection introduces the experimental configurations and comparison performance of our proposed EML model in multi-shot scenario.

\subsubsection{Experimental Configurations} 
In multi-shot scenario, we set $M=2$, \emph{i.e.}, two-shot scenario with three stages for illustration, as depicted in Fig.~\ref{fig:multi-shot}. The streaming samples used in one-shot scenario are split into three stages. 
Except for the configurations introduced in one-shot scenario, the additional experimental configurations for multi-shot scenario are summarized as follows: 
\begin{itemize}
	\setlength{\itemsep}{1pt}
	\setlength{\parsep}{0pt}
	\setlength{\parskip}{0pt}
	\item All batches of T-stage in one-shot scenario are split into Stage 1 and Stage 2 with equal number of samples, as shown in Fig.~\ref{fig:multi-shot}. Under this setting, the survived features in Stage 2 would be the vanished features in Stage 3, and the new augmented features in Stage 2 would be the survived features in Stage 3. In other words, Stage 1 and Stage 2 are respectively considered as T-stage and I-stage for the first feature evolution. Moreover, Stage 2 and Stage 3 are regarded as T-stage and I-stage for the second feature evolution.
	
	\item The features are divided into four equal parts with the same partition order as one-shot scenario. Concretely, the second quarter is the shared part of Stage 1 and Stage 2. The third quarter is the shared part of Stage 2 and Stage 3. The first quarter in Stage 1 and the last quarter in Stage 3 denote the vanished and new augmented features. 
\end{itemize}

\subsubsection{Experiments for Task I and II}
To address the Task I in multi-shot case, we directly use the last two adjacent evolution stages and regard it as the one-shot scenario for predictions, since the streaming data in any two adjacent stages share the common features. To be specific, we first utilize the transforming strategy in Eq.~\eqref{eq:objective_T-stage} on the streaming data in Stage 2 to learn the discriminative distance matrix, and the inheriting strategy in Eq.~\eqref{eq:objective_I-stage} is then applied to classify samples in Stage 3. To tackle the Task II in multi-shot scenario, we regard two adjacent stages as one-shot scenario (\emph{i.e.}, T-stage and I-stage) and repeat this procedure until the last stage in multi-shot scenario. Specifically, the transforming and inheriting strategies are first integrated into Stage 1 and Stage 2, and then we make predictions on the second batch streaming data in Stage 2. After inheriting the metric performance of Stage 1, we extract the useful information from the survived features in Stage 2 and  forward it into the new augmented features via common discriminative space, when new labeled streaming data in Stage 2 arriving. Furthermore, we perform the same inheriting strategy on survived features in Stage 2 to promote the performance predictions in Stage 3.

The experimental results of our proposed EML model averaged over fifty random repetitions for Task I and II are presented in Table~\ref{table:results_multi_shot_task_1} and Fig.~\ref{fig:results_multi_shot_task_2}. Notice that: 1) Our model significantly outperforms other competing methods (\emph{e.g.}, OPIDe and OPID \cite{Hou2018OnePassLW}) especially in Task I, since it could inherit the metric performance of survived features in any two adjacent stages. 2) Compared with Task I, our proposed model performs better for Task II in most cases, since the survived features existing in Stage 1 could effectively promote the predictions for following streaming batches. 3) Our model could be successfully extended from one-shot case into multi-shot scenario to address both Task I and Task II, which further verifies the superior performance of our EML model.

\subsubsection{Ablation Studies}
In this subsection, we conduct extensive variant experiments on Task I and Task II to investigate the efficiency of each component of our EML model in the multi-shot scenario, as introduced in Table~\ref{table:ablation_studies_multi_shot_task_1} and Table~\ref{table:ablation_studies_multi_shot_task_2}. We have the following conclusions from the presented results: 1) All designed components in our EML model could cooperate well to achieve the best performance for both Task I and Task II in the multi-shout scenario, which validates the effectiveness and necessity of each module. 2) Two complementary strategies (\emph{i.e.}, T-stage and I-stage) effectively compress the important information from vanished features and inherit the metric performance from the previous stage. They play an indispensable role in addressing both feature and instance evolutions simultaneously under the Wasserstein distance metric. 3) The performance degradation of Ours-woW illustrates the effectiveness of the smoothed Wasserstein distance to explore the similarity relationships for heterogeneous samples among different stages.

\begin{table}[t]
	\centering
	\setlength{\tabcolsep}{3.0mm}
	\caption{Ablation studies of our proposed EML model in multi-shot scenario for Task I.}
	\scalebox{0.83}{
		\begin{tabular}{|c|c|ccc|c|}
			\hline
			Dataset & $n_i$ & Ours-woT & Ours-woI & Ours-woW & Ours \\
			\hline	
			Mnist    & 80  & 94.93$\pm$0.34 & 94.26$\pm$0.38 & 96.04$\pm$0.62 & \textbf{98.54$\pm$1.08} \\
			0vs5     & 160 & 94.17$\pm$0.55 & 93.41$\pm$0.82 & 95.52$\pm$0.39 & \textbf{98.61$\pm$0.57} \\
			& 320 & 96.13$\pm$0.59 & 95.47$\pm$0.85 & 96.84$\pm$1.03 & \textbf{98.73$\pm$0.64} \\
			
			\hline
			& 100 & 92.45$\pm$0.83 & 91.17$\pm$0.76 & 93.61$\pm$0.35 & \textbf{96.12$\pm$1.18} \\
			Gisette  & 200 & 92.84$\pm$0.72 & 92.04$\pm$0.28 & 94.12$\pm$0.46 & \textbf{95.94$\pm$1.72} \\
			& 300 & 93.05$\pm$0.80 & 92.36$\pm$0.73 & 93.88$\pm$1.14 & \textbf{95.71$\pm$1.68} \\
			
			\hline
			USPS     & 120 & 95.54$\pm$0.75 & 94.18$\pm$0.93 & 96.33$\pm$0.41 & \textbf{98.57$\pm$0.94} \\
			0vs5     & 160 & 95.06$\pm$0.83 & 94.27$\pm$0.53 & 96.84$\pm$0.65 & \textbf{98.68$\pm$0.65} \\
			& 240 & 95.36$\pm$0.32 & 94.91$\pm$0.77 & 97.05$\pm$0.41 & \textbf{98.39$\pm$0.72} \\

			\hline
			USPS     & 180 & 90.15$\pm$0.19 & 89.35$\pm$0.87 & 90.94$\pm$0.51 & \textbf{93.11$\pm$1.87} \\
			0vs3vs5  & 240 & 90.62$\pm$0.30 & 89.87$\pm$0.64 & 91.58$\pm$0.74 & \textbf{93.23$\pm$1.58} \\
			& 300 & 90.86$\pm$0.81 & 90.22$\pm$0.63 & 92.08$\pm$0.26 & \textbf{93.13$\pm$1.55} \\
			\hline
		\end{tabular}
	}
	\label{table:ablation_studies_multi_shot_task_1}
\end{table}

\begin{table}[t]
	\centering
	\setlength{\tabcolsep}{3.0mm}
	\caption{Ablation studies of our proposed EML model in multi-shot scenario for Task II.}
	\scalebox{0.83}{
		\begin{tabular}{|c|c|ccc|c|}
			\hline
			Dataset & $n_i$ & Ours-woT & Ours-woI & Ours-woW & Ours \\
			\hline	
			Mnist    & 80  & 94.58$\pm$1.48 & 93.26$\pm$1.71 & 95.93$\pm$1.22 & \textbf{98.30$\pm$1.18} \\
			0vs5     & 160 & 95.71$\pm$1.29 & 93.62$\pm$1.48 & 96.04$\pm$0.98
			& \textbf{98.13$\pm$1.16} \\
			& 320 & 95.25$\pm$0.93 & 94.54$\pm$0.89 & 95.87$\pm$1.13
			& \textbf{98.24$\pm$1.34} \\
			
			\hline
			& 100 & 94.32$\pm$0.88 & 93.68$\pm$1.09 & 95.02$\pm$0.95
			& \textbf{97.22$\pm$1.37} \\
			Gisette  & 200 & 93.10$\pm$1.27 & 92.29$\pm$1.48 & 94.21$\pm$1.07
			& \textbf{96.90$\pm$1.18} \\
			& 300 & 93.74$\pm$1.18 & 92.16$\pm$1.26 & 94.23$\pm$0.89
			& \textbf{96.92$\pm$1.70} \\
			
			\hline
			USPS     & 120 & 96.15$\pm$1.36 & 95.62$\pm$1.26 & 96.87$\pm$0.88
			& \textbf{98.17$\pm$1.18} \\
			0vs5     & 160 & 94.77$\pm$1.38 & 94.46$\pm$1.52 & 95.45$\pm$1.13
			& \textbf{98.14$\pm$0.97} \\
			& 240 & 95.32$\pm$1.36 & 94.37$\pm$1.65 & 96.14$\pm$0.76
			& \textbf{98.67$\pm$0.62} \\

			\hline
			USPS     & 180 & 90.38$\pm$1.47 & 90.56$\pm$1.51 & 91.06$\pm$1.05
			& \textbf{93.94$\pm$1.50} \\
			0vs3vs5  & 240 & 90.43$\pm$1.26 & 89.85$\pm$1.34 & 91.88$\pm$1.17
			& \textbf{93.34$\pm$1.38} \\
			& 300 & 90.35$\pm$1.18 & 90.83$\pm$1.43 & 91.46$\pm$0.84
			& \textbf{94.58$\pm$1.25} \\
			\hline
		\end{tabular}
	}
	\label{table:ablation_studies_multi_shot_task_2}
\end{table}

\section{Conclusion}
In this paper, an online Evolving Metric Learning (EML) model is proposed for both instance and feature evolutions, which is successfully applied to one-shot and multi-shot scenarios. Our proposed EML model contains two essential stages, \emph{i.e.}, Transforming stage (T-stage) and Inheriting stage (I-stage). To be specific, for the T-stage, we utilize the survived features to characterize the effective information extracted from vanished and survived features by exploiting a common discriminative metric space. In the I-stage, we inherit the metric performance of survived features from T-stage, and extend it into the new augmented features. Furthermore, we apply the smoothed Wasserstein distance to T-stage and I-stage to better explore the similarity relations of heterogeneous streaming data among different evolution stages. Extensive experiments show the superior performance of our proposed EML model on several representative datasets. In the future, we will consider lifelong machine learning for both instance and feature evolutions, which continually learns a sequence of new streaming evolution tasks without the catastrophic forgetting for the previous learned evolution tasks.

\ifCLASSOPTIONcaptionsoff
  \newpage
\fi

\bibliographystyle{IEEEtran}
\bibliography{FeatureEvolution}

\begin{IEEEbiography}[{\includegraphics[width=1in,height=1.25in,clip,keepaspectratio]{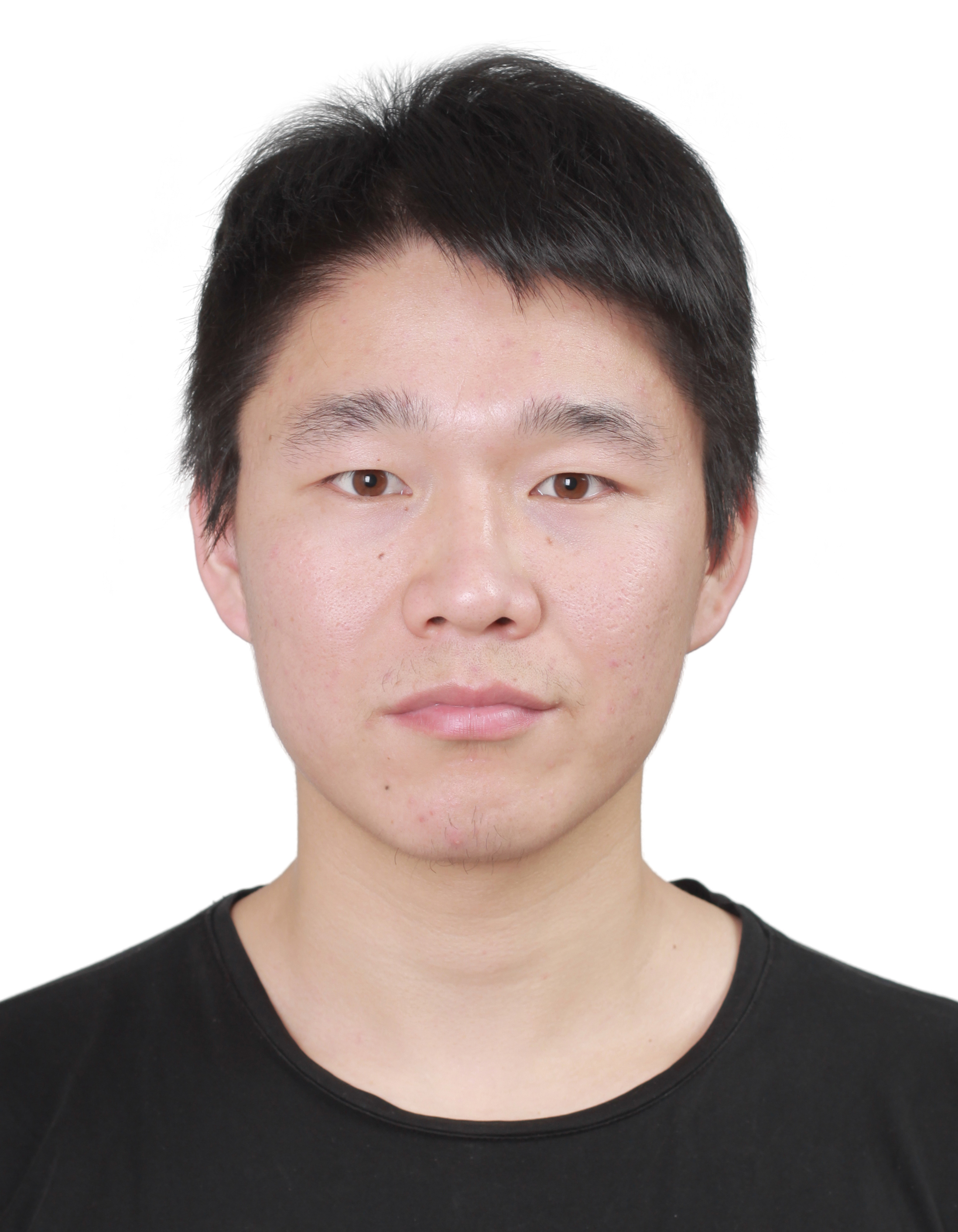}}]{Jiahua Dong} Jiahua Dong is currently a Ph. D candidate in State Key Laboratory of Robotics, Shenyang Institute of Automation, Chinese Academy of Sciences. He received the B.S. degree from Jilin University in 2017. His current research interests include computer vision, machine learning, transfer learning, domain adaptation and medical image processing.
\end{IEEEbiography}

\begin{IEEEbiography}[{\includegraphics[width=1in,height=1.25in,clip,keepaspectratio]{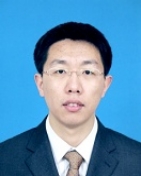}}]{Yang Cong} Yang Cong (S’09-M’11-SM’15) is a full professor of Chinese Academy of Sciences. He received the he B.Sc. de. degree from Northeast University in 2004, and the Ph.D. degree from State Key Laboratory of Robotics, Chinese Academy of Sciences in 2009. He was a Research Fellow of National University of Singapore (NUS) and Nanyang Technological University (NTU) from 2009 to 2011, respectively; and a visiting scholar of University of Rochester. He has served on the editorial board of the Journal of Multimedia. His current research interests include image processing, compute vision, machine learning, multimedia, medical imaging, data mining and robot navigation. He has authored over 70 technical papers. He is also a senior member of IEEE.
\end{IEEEbiography}

\begin{IEEEbiography}[{\includegraphics[width=1in,height=1.25in,clip,keepaspectratio]{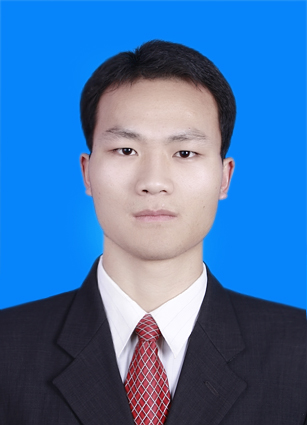}}]{Gan Sun} Gan Sun (S'19) is an Assistant Professor in State Key Laboratory of Robotics, Shenyang Institute of Automation, Chinese Academy of Sciences. He received the B.S. degree from Shandong Agricultural University in 2013, the Ph.D. degree from State Key Laboratory of Robotics, Chinese Academy of Sciences in 2020, and has been visiting Northeastern University from April 2018 to May 2019, Massachusetts Institute of Technology from June 2019 to November 2019. His current research interests include lifelong machine learning, multi-task learning, medical data analysis, deep learning and 3D computer vision.
\end{IEEEbiography}

\begin{IEEEbiography}[{\includegraphics[width=1in,height=1.25in,clip,keepaspectratio]{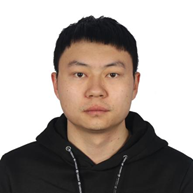}}]{Tao Zhang} Tao Zhang is currently working toward the Ph.D. degree in pattern recognition and intelligent systems at the State Key Laboratory of Robotics, Shenyang Institute of Automation, Chinese Academy of Sciences, Shenyang, China. His research interests include pattern recognition, image processing, tactile sensing and robotics.
\end{IEEEbiography}

\begin{IEEEbiography}[{\includegraphics[width=1in,height=1.25in,clip,keepaspectratio]{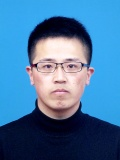}}]{Xu Tang} Xu Tang is currently a reserach associate in State Key Laboratory of Robotics, Shenyang Institute of Automation. He received MESc degree from Harbin Institute of Technology in 2017. His current research interests include computer vision and machine learning.
\end{IEEEbiography}

\begin{IEEEbiography}[{\includegraphics[width=1in,height=1.25in,clip,keepaspectratio]{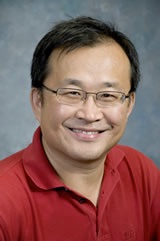}}]{Xiaowei Xu} Xiaowei Xu is a professor of Information Science at University of Arkansas at Little Rock (UALR), received a  a B.Sc. de. degree in Mathematics from Nankai University in 1983 and a Ph.D. degree in Computer Science from University of Munich in 1998. He holds an adjunct professor position in the Department of Mathematics and Statistics at University of Arkansas at Fayetteville. Before his appointment in UALR, he was a senior research scientist in Siemens. He was a visiting professor in Microsoft Research Asia and Chinese University of Hong Kong. His research spans data mining, machine learning, bioinformatics, data management and high performance computing. He has published over 70 papers in peer reviewed journals and conference proceedings. His groundbreaking work on density-based clustering algorithm DBSCAN has been widely used in many textbooks; and received over 10203 citations based on Google scholar. Dr. Xu is a recipient of 2014 ACM KDD Test of Time Award that “recognizes outstanding papers from past KDD Conferences beyond the last decade that have had an important impact on the data mining research community.”
\end{IEEEbiography}

\end{document}